\documentclass{article}

\usepackage{graphicx}
\usepackage{epstopdf}
\usepackage{subfigure}
\usepackage{hyperref}
\usepackage[hyphenbreaks]{breakurl}
\PassOptionsToPackage{hyphens}{url}
\usepackage{array}
\usepackage{makecell}
\usepackage{algorithm}
\usepackage{amsmath}
\usepackage{booktabs}
\usepackage{verbatim}
\usepackage{tabularx}
\usepackage{algpseudocode}
\usepackage{amssymb}

\usepackage[table,xcdraw]{xcolor}

\usepackage[nonatbib]{neurips_2023}

\usepackage[utf8]{inputenc} % allow utf-8 input
\usepackage[T1]{fontenc}    % use 8-bit T1 fonts
\usepackage{hyperref}       % hyperlinks
\usepackage{url}            % simple URL typesetting
\usepackage{booktabs}       % professional-quality tables
\usepackage{amsfonts}       % blackboard math symbols
\usepackage{nicefrac}       % compact symbols for 1/2, etc.
\usepackage{microtype}      % microtypography
\usepackage{xcolor}         % colors
\usepackage{listings}
\usepackage{float}
\definecolor{codegreen}{rgb}{0,0.6,0}
\definecolor{codegray}{rgb}{0.5,0.5,0.5}
\definecolor{codepurple}{rgb}{0.58,0,0.82}
\definecolor{backcolour}{rgb}{0.95,0.95,0.92}

\lstdefinestyle{mystyle}{
    backgroundcolor=\color{backcolour},   
    commentstyle=\color{codegreen},
    keywordstyle=\color{magenta},
    numberstyle=\tiny\color{codegray},
    stringstyle=\color{codepurple},
    basicstyle=\ttfamily\footnotesize,
    breakatwhitespace=false,         
    breaklines=true,                 
    captionpos=b,                    
    keepspaces=true,                 
    numbers=left,                    
    numbersep=5pt,                  
    showspaces=false,                
    showstringspaces=false,
    showtabs=false,                  
    tabsize=2
}

\title{MIRACLE: Multi-Task Learning Based Interpretable Regulation of Autoimmune Diseases through Common Latent Epigenetics}

\author{%
  Pengcheng Xu \footnotemark[1]\\
  Department of Electrical and Computer Engineering\\
  University of Illinois Urbana Champaign\\
  Urbana,IL,61801 \\
  \texttt{px6@illinois.edu} \\
   \And
   Jinpu Cai \footnotemark[1]\\
  University of Michigan-Shanghai Jiao Tong University Joint Institute \\
  Shanghai Jiao Tong University\\
  Shanghai,China \\
   \texttt{jinpu.cai@sjtu.edu.cn} \\
   \And
   Yulin Gao \\
   Department of Statistics,The College of Literature, Science and the Arts \\
  University of Michigan\\
  Ann Arbor, MI, 48105\\
   \texttt{yulingao@umich.edu} \\
   \And
   Ziqi Rong \\
   School of Information \\
  University of Michigan\\
  Ann Arbor, MI, 48105 \\
   \texttt{ziqirong@umich.edu} \\
  \AND
  \footnotetext[1]{These authors contributed equally to this work.} %对应脚注[1]
  \footnotetext[2]{Corresponding authors.} %对应脚注[2]
}

\begin{document}

\maketitle

\begin{abstract}
DNA methylation is a crucial regulator of gene transcription and has been linked to various diseases, including autoimmune diseases and cancers. However, diagnostics based on DNA methylation face challenges due to large feature sets and small sample sizes, resulting in overfitting and suboptimal performance. To address these issues, we propose MIRACLE, a novel interpretable neural network that leverages autoencoder-based multi-task learning to integrate multiple datasets and jointly identify common patterns in DNA methylation.

MIRACLE's architecture reflects the relationships between methylation sites, genes, and pathways, ensuring biological interpretability and meaningfulness. The network comprises an encoder and a decoder, with a bottleneck layer representing pathway information as the basic unit of heredity. Customized defined MaskedLinear Layer is constrained by site-gene-pathway graph adjacency matrix information, which provides explainability and expresses the site-gene-pathway hierarchical structure explicitly. And from the embedding, there are different multi-task classifiers to predict diseases.

Tested on six datasets, including rheumatoid arthritis, systemic lupus erythematosus, multiple sclerosis, inflammatory bowel disease, psoriasis, and type 1 diabetes, MIRACLE demonstrates robust performance in identifying common functions of DNA methylation across different phenotypes, with higher accuracy in prediction dieseases than baseline methods. By incorporating biological prior knowledge, MIRACLE offers a meaningful and interpretable framework for DNA methylation data analysis in the context of autoimmune diseases.
\end{abstract}

\section{Introduction}

Epigenomics sequencing is a technology that studies gene expression and regulation of heritable changes without altering DNA sequences\cite{moore2013dna}. DNA methylation represents a direct chemical modification of DNA. Specifically, methylation at the fifth position of cytosine (5mc) has been identified as a stable epigenetic marker\cite{kader2020characterization}. Alterations in DNA methylation patterns have been documented in a variety of human diseases, including cancer \cite{das2004dna} and autoimmune disorders\cite{kim2017dna,funes2021contribution}. 

The mechanism that maintains DNA methylation patterns is sensitive to both exogenous and endogenous influences\cite{richardson2003dna}. Research has demonstrated that the administration of exogenous DNA methylation inhibitors can induce lupus-like symptoms in murine models\cite{quddus1993treating}. Additionally, endogenous retroviruses (HERVs) have been shown to contribute to the development of lupus by inhibiting the methylation of long terminal repeat sequences (LTRS) in CD4 positive T cells\cite{sun2016dna}. Autoimmune diseases are characterized by overlapping features, a phenomenon first described by Noel R Rose as "shared threads"\cite{rose1997autoimmune}. Genome-wide association studies (GWAS) and the identification of risk-associated single-nucleotide polymorphisms (SNPs) have provided evidence that the genetic basis of autoimmune diseases may involve multiple loci encoding low-risk susceptibility alleles\cite{mackay2009clustering}. These loci have been implicated in multiple autoimmune diseases, suggesting the existence of common regulatory mechanisms underlying their expression.

Elucidating the relationship between DNA methylation and disease, as well as the underlying biological regulatory mechanisms, remains a formidable challenge in the field. Firstly. The core challenges are as follows: Array-based methods such as the Infinium Human Methylation 450K have been widely employed for the quantitative analysis of methylation sites in cancer and autoimmune diseases\cite{haider2020dna}. However, the interpretation of such data is complicated by high dimensionality and low sample sizes \cite{gatev2020comeback}. To overcome these challenges, researchers have developed a range of machine learning models that can handle high-dimensional data.Despite these advances, machine learning models are still prone to overfitting, which can result in a high rate of errors in disease diagnosis.

% \item[2. Interpretability] Secondly. Current approaches to methylation biomarker detection primarily focus on single diseases and do not take into account the interpretability of high-dimensional methylation features. 

% \item[3. Commonalities in between autoimmune disease are not harnessed] All current methods are aimed at a single disease task, without considering the commonality of associations of a single type of disease. 
% \end{description}

Current approaches to methylation biomarker detection primarily focus on single diseases and do not take into account the interpretability of high-dimensional methylation features. ReGear uses linear models to associate methylation sites with genes in order to elucidate the regulatory mechanisms underlying methylation\cite{cai2021comprehensive}. However, ReGear only consider linear correlations and do not account for the impact of genes on biological function. Despite their utility, current methods for methylation biomarker detection are limited in their ability to account for the impact of genes on biological function. These methods typically rely on linear correlations and do not consider the commonalities among different diseases.

To address the aforementioned challenges in elucidating the relationship between DNA methylation and disease, as well as the underlying biological regulatory mechanisms, we developed MIRACLE. This novel approach leverages multi-task learning and advanced deep learning techniques to overcome limitations associated with high-dimensional features, low-dimensional samples, interpretability, and single-disease focus.

MIRACLE incorporates multi-task learning, an approach that improves performance by leveraging shared features across multiple related tasks\cite{zhang2021survey}. By employing multi-task learning in the context of DNA methylation data analysis, MIRACLE addresses the first challenge of high-dimensionality. This technique reveals common functions of DNA methylation in different phenotypes, which could provide valuable insights for biological research. Meanwhile, our method acknowledges that current deep learning models often lack interpretability. To remedy this, we construct a residue-gene-pathway network using an interpretable network, which includes a variational autoencoder (VAE) and a MaskedLinear Layer\cite{vahdat2020nvae}. This interpretable network not only conforms to biological laws but also provides a better understanding of the regulatory mechanisms underlying DNA methylation.

In summary, MIRACLE demonstrates its versatility and applicability across a wide landscape of autoimmune diseases by achieving the following objectives:Firstly, it attains high accuracy in predicting multiple autoimmune diseases based on DNA methylation data.Secondly, it uncovers common functions of DNA methylation in various phenotypes through multi-task learning.Thirdly, it inherently offers interpretability, thanks to the MaskedLinear Layer of VAE, whose three layers are defined by the biology ontology. The ontology establishes connections between sites, genes, and pathways.
% \begin{itemize}
% \item[$\bullet$]Firstly, it attains high accuracy in predicting multiple autoimmune diseases based on DNA methylation data.
% \item[$\bullet$]Secondly, it uncovers common functions of DNA methylation in various phenotypes through multi-task learning.
% \item[$\bullet$]Thirdly, it inherently offers interpretability, thanks to the MaskedLinear Layer of VAE, whose three layers are defined by the biology ontology. The ontology establishes connections between sites, genes, and pathways.
% \end{itemize}

\section{Related Works}
% \subsection{Traditional Machine Learning Methods}
% To overcome the challenge of high dimension feature and low dimension sample in the DNA methylation data, researchers have developed a range of machine learning models that can handle high-dimensional data.
Previous studies mainly focused on sample category prediction based on a single disease dataset. Due to the small number of samples, neural networks are difficult to train in a single dataset. Therefore, current research is almost entirely based on traditional machine learning algorithms. This paper compares 10 machine learning algorithms, including regression-based linear regression\cite{mcewen2018systematic} and logistics regression\cite{wei2021cpgtools}. At the same time, Bayesian classification has also been used in DNA methylation research\cite{zhang2015predicting,ramakrishnan2016analysis}. Boosting proposes more general models by reducing the bias of supervised learning\cite{mahendran2022deep,li2018human}. Random forests train samples by constructing multiple decision trees\cite{zhang2015predicting}. K-nearest neighbors predict sample categories based on the Euclidean distance between samples\cite{ma2022diagnostic}.

\section{Methods}

\subsection{Model architecture}
\subsubsection{Overview}

\begin{figure}
        \centering
        \includegraphics[width=1.05\linewidth]{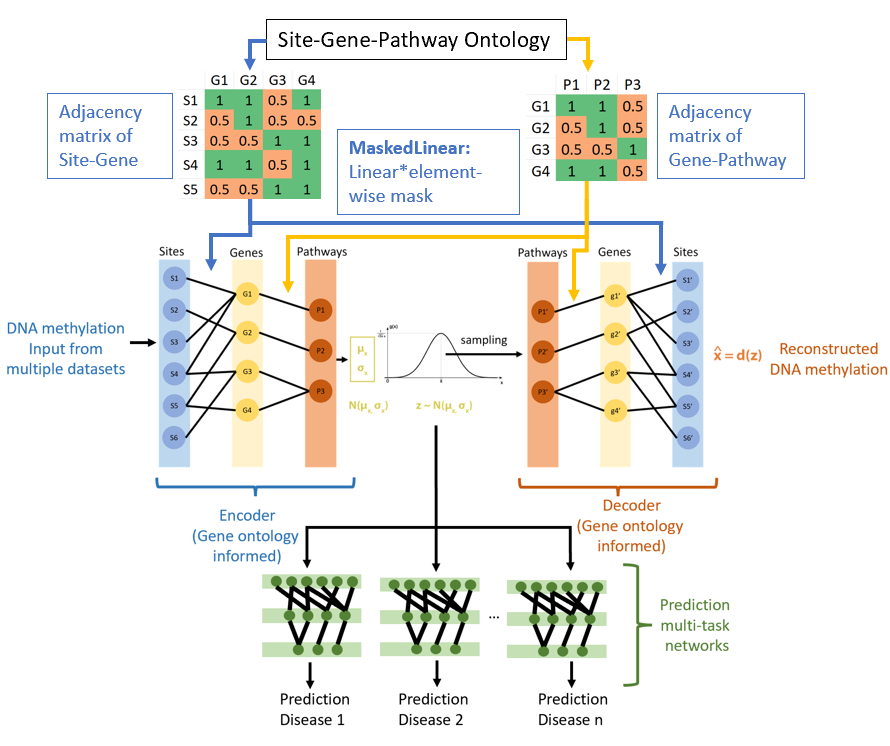}%MeiNN-new.png%MeiNN-new.png
   \caption{MIRACLE:An interpretable variational auto-encoder based on multi-task NN}
\end{figure}

We first used t-test to filter out noise features and reduce the complexity of neural networks by selecting top significantly different (smaller p-value) features\cite{model2001feature}. Details of pre-processing can be found in appendix.

The MIRACLE network consists of an encoder and a decoder and n classifiers(n depends on number of tasks). The encoder distills the critical and low-dimension embedding data from residual methylation without human labeling, while the decoder is designed to be explainable by pruning nodes and restoring data from pathway to gene-level methylation and residual methylation. The network is trained on multiple datasets, including rheumatoid arthritis, systemic lupus erythematosus, multiple sclerosis, and inflammatory bowel disease, to identify common functions of DNA methylation in phenotypes.

\textbf{The autoencoder} Its input is site methylation and the output is reconstructed site methylation.

\textbf{The encoder} Its input is site methylation, and its output is embedding. It have Masked Linear which is defined by prior biology knowledge of "site-gene-pathway" connection. The number of nodes in each layer of the encoder follows "site-gene-pathway" dimension.This structure follows a hierarchical ontology based on prior biological knowledge\cite{kanehisa2000kegg} in public data. Each layer forward has smaller dimension as the previous layer, creating a bottleneck layer that represents the pathway and provides information about the basic units of heredity.

\textbf{The decoder} Its input is embedding, and output is reconstructed site methylation. It has Masked Linear and reconstruct data from pathway to gene-level methylation and then to site methylation. The sparse network reduces the dimension to compute and can be explained by gene rules.

\textbf{The classifiers} are fully connected neural networks for specific disease prediction tasks. Here we have rheumatoid arthritis, systemic lupus erythematosus, multiple sclerosis, and inflammatory bowel disease, psoriasis, and type 1 diabetes.

The whole network is trained together and is assumed to identify common features in the embedding layer and predict different diseases.
 
\begin{table}[h]
\centering
\caption{Notations}
\begin{tabular}{cl}
\toprule
Symbol & Description \\
\midrule
$n$ & Dimension of selected sites ($\mathbb{R}$) \\
$g$ & Dimension of genes in the biology ontology ($\mathbb{R}$) \\
$p$ & Dimension of pathways in the biology ontology ($\mathbb{R}$) \\
$d$ & Number of samples ($\mathbb{R}$) \\
$t$ & Number of tasks ($\mathbb{R}$) \\
$\mathbf{x}$ & $\mathbb{R}^{n\times d}$ \\
$\mathbf{Y}$ & $\mathbb{R}^{t\times d}$ \\
$Enc$ & Encoder: $\mathbb{R}^n \rightarrow \mathbb{R}^p$ \\
$Dec$ & Decoder: $\mathbb{R}^p \rightarrow \mathbb{R}^n$ \\
$Cla_i$ & $n$ multi-task classifiers: $\mathbb{R}^p \rightarrow \mathbb{R}^1$ \\

\bottomrule
\end{tabular}
\end{table}
\textbf{Loss function}: ($\alpha,\beta,\gamma_i$ is hyper-parameter of each weight of each partial loss)
$$
Loss= \alpha \times MSE(\mathbf{x},Dec \circ Enc(\mathbf{x}))+\beta \times D_{KL}(Enc(\mathbf{x})) + \sum^t_{i=1}{\gamma_i\times BCE(Cla_{i} \circ Enc(\mathbf{x}),\mathbf{Y})}
$$
Note that in each stage or epoch, the hyper-parameters $\alpha,\beta,\gamma_i$ may change according to some specific optimization algorithms. \\
MSE means Mean Square Error, BCE means Binary Cross-Entropy.$D_{KL}$ means KL-divergence.\\

\subsubsection{Autoencoder}
Autoencoders, first introduced in [5], are a form of neural network trained to reconstruct input data. The primary goal is unsupervised learning of significant data representations, useful in applications such as clustering. The defined problem [6] is to ascertain functions $A: \mathbb{R}^n \rightarrow \mathbb{R}^p$ (encoder) and $B: \mathbb{R}^p \rightarrow \mathbb{R}^n$ (decoder) meeting the condition:
$$
\arg \min _{A, B} E[\Delta(\mathbf{x}, B \circ A(\mathbf{x})]
$$
where $E$ denotes the expectation over $x$ distribution, and $\Delta$ is the reconstruction loss function, typically the $\ell_2$-norm, quantifying the distance between the decoder's output and its input.

Autoencoders present a pivotal bias-variance tradeoff. While the architecture aims for accurate input reconstruction, reducing the reconstruction error, it also seeks a low, generalizable representation.

\textbf{Sparse autoencoders}

To navigate the bias-variance tradeoff, sparsity enforcement on hidden activations can be implemented, either complementing or substituting bottleneck enforcement. Two strategies exist for sparsity regularization, akin to standard regularization, but applied to activations instead of weights. The first strategy involves $L_1$ regularization, which promotes sparsity, modifying the autoencoder optimization objective to:
$$
\arg \min _{A, B} E[\Delta(\mathbf{x}, B \circ A(\mathbf{x})]+\lambda \sum_i\left|a_i\right|,
$$
where $a_i$ is the activation of the $i$th hidden layer, with $i$ iterating across all hidden activations.

\textbf{Variational autoencoders}

Variational Autoencoders (VAE), a considerable advancement in autoencoders' representational abilities, are generative models inspired by Variational Bayes (VB) Inference. VAEs attempt to model data generation via a probabilistic distribution. Specifically, given an observed dataset  $\mathbf{X}=\left\{\mathbf{x}_i\right\}_{i=1}^N$  comprising $V$ i.i.d samples, we propose a generative model for each instance $\mathbf{x}_i$ contingent on an unobserved random latent variable $\mathbf{z}_i$, governed by parameters $\theta$. This generative model also functions as a probabilistic decoder. In parallel, we propose an approximate posterior distribution over $\mathbf{z}_i$ given $\mathbf{x}_i$, denoted as recognition, acting as a probabilistic encoder, governed by parameters $\phi$. We also propose a prior distribution for latent variables $\mathbf{z}i$, denoted by $p\theta\left(\mathbf{z}i\right)$. These latent variables, observable as codes from the recognition model $q\phi(\mathbf{z} \mid \mathbf{x})$, are governed by unknown parameters $\theta$ and $\phi$ to be learned from data.

The marginal log-likelihood is the sum over individual data points $\log p_\theta\left(\mathbf{x}_1, \mathbf{x}2, \ldots, \mathbf{x}N\right)=\sum{i=1}^N \log p\theta\left(\mathbf{x}_i\right)$, each of which can be expressed as:
$$
\log p_\theta\left(\mathbf{x}_i\right)=D_{K L}\left(q_\phi\left(\mathbf{z} \mid \mathbf{x}_i\right) \| p_\theta\left(\mathbf{z} \mid \mathbf{x}_i\right)\right)+\mathcal{L}\left(\theta, \phi ; \mathbf{x}_i\right)
$$

where the first term is the Kullback-Leibler divergence of the approximate recognition model from the true posterior and the second term, the variational lower bound on the marginal likelihood, is defined as:
$$
\mathcal{L}\left(\theta, \phi ; \mathbf{x}_i\right) \triangleq \mathbb{E}_{q_\phi\left(\mathbf{z} \mid \mathbf{x}_i\right)}\left[-\log q_\phi(\mathbf{z} \mid \mathbf{x})+\log p_\theta(\mathbf{x}, \mathbf{z})\right]
$$

As the Kullback-Leibler divergence is non-negative, $\mathcal{L}\left(\theta, \phi ; \mathbf{x}_i\right)$ forms a lower bound on the marginal log-likelihood. Maximizing this lower bound optimizes our approximation of the posterior in terms of Kullback-Leibler divergence, given that the marginal log-likelihood is independent of $\theta$ and $\phi$.

Further, the variational lower bound can be expanded as:
$$
\mathcal{L}\left(\theta, \phi ; \mathbf{x}_i\right)=-D_{K L}\left(q_\phi\left(\mathbf{z} \mid \mathbf{x}_i\right) \| p_\theta(\mathbf{z})\right)+\mathbb{E}_{q_\phi\left(\mathbf{z} \mid \mathbf{x}_i\right)}\left[\log p_\theta\left(\mathbf{x}_i \mid \mathbf{z}\right)\right]
$$
Variational inference is achieved by maximizing $\mathcal{L}\left(\theta, \phi ; \mathbf{x}_i\right)$ for all instances concerning $\theta$ and $\phi$.

Given dataset $\mathbf{X}=\left\{\mathbf{x}_i\right\}_{i=1}^N$  with $N$ instances, the marginal likelihood lower-bound of the complete dataset $\mathcal{L}(\theta, \phi ; \mathbf{X})$ can be estimated using a mini-batch $\mathbf{X}^M=$ $\left\{\mathbf{x}_i\right\}_{i=1}^M$ of size $M$ as:

$$
\mathcal{L}(\theta, \phi ; \mathbf{X}) \approx \tilde{\mathcal{L}}^M\left(\theta, \phi ; \mathbf{X}^M\right)=\frac{N}{M} \sum_{i=1}^M \mathcal{L}\left(\theta, \phi ; \mathbf{x}_i\right)
$$
While classical mean-field VB presumes a factorized approximate posterior followed by closed-form optimization updates (usually requiring conjugate priors), VAEs adopt a distinct approach, approximating gradients of $\tilde{\mathcal{L}}^M\left(\theta, \phi ; \mathbf{X}^M\right)$ using the reparameterization trick and stochastic gradient optimization.

\subsubsection{MaskedLinear layer: Site-Gene-Pathway ontology prior constraint to neural network}

The MaskedLinear layer, leveraged for the interpretability of neural networks with biological graph knowledge (site-gene-pathway relations), is defined mathematically with an input tensor $X \in \mathbb{R}^{m \times n}$ and a masked weight matrix $W^{masked} \in \mathbb{R}^{n \times p}$, obtained by element-wise multiplication of the original weight matrix $W$ and the mask matrix $M$. The output tensor $Y \in \mathbb{R}^{m \times p}$ is calculated as $Y = XW^{masked} + B$.

Adjacency matrices, $A_{site-gene} \in \mathbb{R}^{n \times q}$ and $A_{gene-pathway} \in \mathbb{R}^{q \times p}$, represent the site-gene and gene-pathway connections respectively, and serve as masks in the MaskedLinear layer. These masks are applied in sequence to the first and second MaskedLinear layers $ML_1$ and $ML_2$. Output is computed via:

\begin{align}
X_{gene} &= X_{site} (W_1 \odot A_{site-gene}) + B_1 \\
Y_{pathway} &= X_{gene} (W_2 \odot A_{gene-pathway}) + B_2
\end{align}

Using adjacency matrices as masks ensures that the model only considers known biological connections, improving model interpretability and facilitating better understanding of the biological system.

To enhance the representation ability of MaskedLinear, we introduce modified adjacency matrices $\tilde{A}{site-gene} \in \mathbb{R}^{n \times q}$ and $\tilde{A}{gene-pathway} \in \mathbb{R}^{q \times p}$, where values indicate connection strengths and range within $[0, 1]$. These matrices are used as masks in the MaskedLinear layers, and output is computed via:

\begin{align}
X_{gene} &= X_{site} (W_1 \odot \tilde{A}{site-gene}) + B_1 \\
Y{pathway} &= X_{gene} (W_2 \odot \tilde{A}_{gene-pathway}) + B_2
\end{align}

This adjustment allows for flexible modeling of connection strengths and potentially enhances the model's representational capacity and accuracy.

%%%%%%%%%%%%%%%%%%%%%%%%%%%%%%%%%%%%%%%%%%%%%%%%%%%%%%%%%%%%%%%%%
\begin{figure}[H]
      \centering
        \includegraphics[width=1\linewidth]{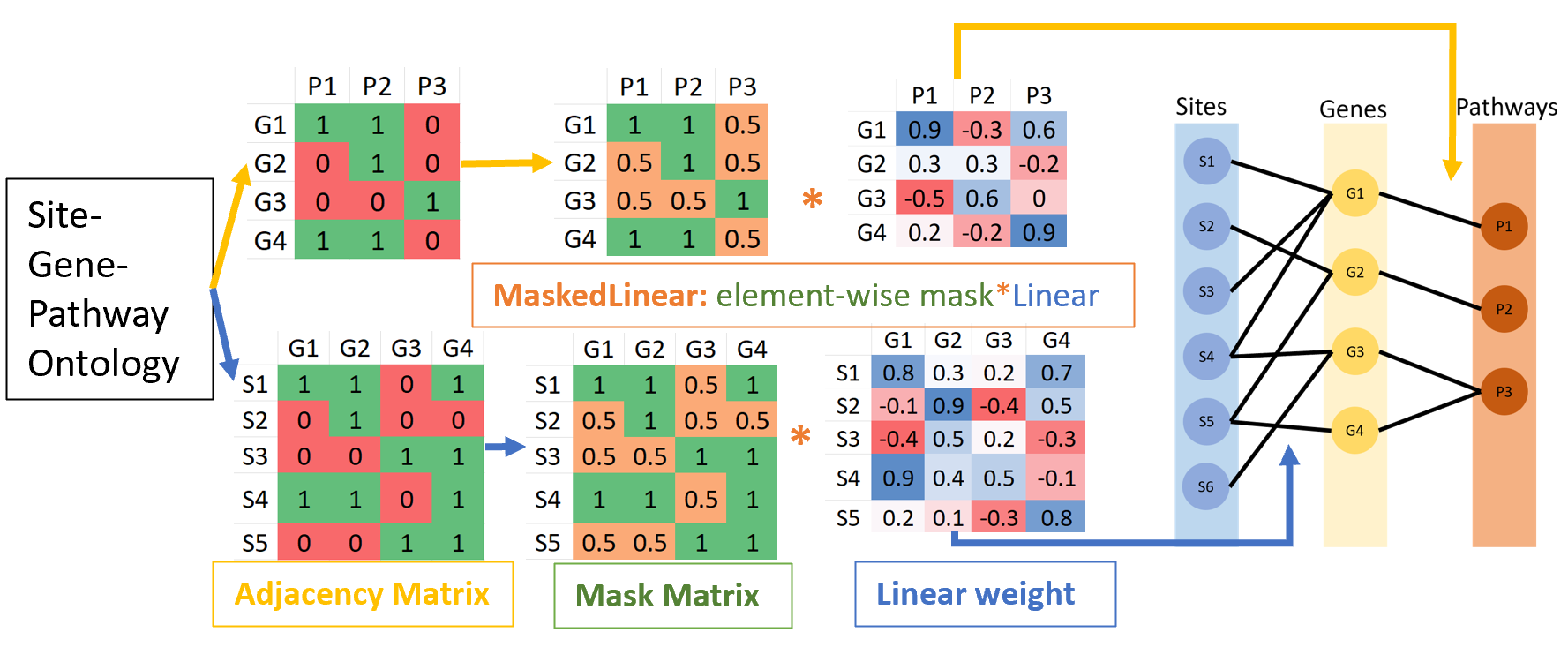}
   \caption{Site-Gene-Pathway ontology adjacency matrix to be MaskLinear Layer}
\end{figure}

We incorporate biology ontology prior knowledge related to methylation site data into the auto-encoder structure, adding a hierarchical site-gene-pathway structure. Our input, representing sites, has a dimension of 480k. Subsequent layers represent genes and pathways, with dimensions around 10k and less than 10k respectively. Connections between neurons in each layer are defined based on the known biology ontology of site-gene-pathway, and not fully-connected. GESA data is utilized as the data source for site-gene-pathway ontology.
% \begin{figure}[H]
% \subfigure
% {
%     \begin{minipage}[b]{0.5\linewidth}
%         \centering
%         \includegraphics[width=1\linewidth]{ontology.png}
%    \caption{Site-Gene-Pathway ontology}
%     \end{minipage}
% }
% \subfigure
% {
%     \begin{minipage}[b]{0.5\linewidth}
%         \centering
%         \includegraphics[width=0.7\linewidth]{site-gene-pathway.png}
%    \caption{Site-Gene-Pathway constrained neural network architecture}
%     \end{minipage}
% }
% \end{figure}
% %%%%%%%%%%%%%%%%%%%%%%%%%%%%%%%%%%%%%%%%%%%%%%%%5
% The structure of the biology ontology is expressed by two adjacency matrices.
% One represents site to gene relationship, the other represents gene to pathway relationship.

% We now have two ways to implement the neural network architecture.
% One way is regularization, which is soft constraint. The other way is mask of neuron, making the neural network sparse, which is hard constraint.
%%%%%%%%%%%%%%%%%%%%%%%%%%%%%%%%%%%%%%%%%%%%%

\subsubsection{Multi-task learning classifier}

We employ a multi-task learning classifier to predict multiple diseases, leveraging the shared knowledge across tasks to improve model performance. The input to the multi-task learning classifier is the embedding generated by the variational autoencoder. In this study, we utilize six classifiers, each connected to the disease output label.

\textbf{Definition of multi-task learning}: Given a set of $m$ related learning tasks $\left\{\mathcal{T}_i\right\}_{i=1}^m$, multi-task learning seeks to jointly learn these tasks, enhancing the model performance for each individual task $\mathcal{T}_i$ by exploiting shared knowledge from other tasks.

\textbf{Training of multi-task learning}:
A collection of $k$ tasks is considered, with each task represented as $D_i^{\operatorname{tr}}=\left\{\left(x_j^{(i)}, y_j^{(i)}\right)\right\}_{j=1}^{n_i}$.
Input data $X_i$ and output data $Y_i$ can be heterogeneous across tasks, indicating they may reside in different spaces (e.g., object classification, object detection, image segmentation, etc.).
Each task can have a distinct loss function, denoted as $\ell_i(\cdot, \cdot)$.
The aim is to achieve better generalization by simultaneously training all tasks, as opposed to training separate models for each individual task.

\textbf{Testing of multi-task learning}:
The same set of tasks is evaluated on test data: $D_i^{\operatorname{te}}=\left\{\left(x_j^{(i)}, y_j^{(i)}\right)\right\}_{j=1}^{n_i}$. The learned model is assessed on each task. A simple baseline for comparison is single-task learning.

\textbf{The multi-task learning (MTL) loss} is defined as: $\min {g, h_i} \frac{1}{k} \sum{i=1}^k \frac{1}{n_i} \sum_{j=1}^{n_i} \ell_i\left(\left(h_i \circ g\right)\left(x_j^{(i)}\right), y_j^{(i)}\right)$, where $g$ denotes the shared knowledge representation, and $h_i$ represents the task-specific functions. The objective is to minimize the average loss across all tasks by optimizing both the shared representation and the task-specific functions.

\subsection{Training scheme and optimization}
\subsubsection{Training scheme}
In order to minimize our loss, we designed a three stage training scheme, which is inspired by the pre-train and fine-tune scheme which is widely adopted in large model area.\\
\textbf{Training stages of model}:\\
1:update weight of auto-encoder and n classifiers\\  
2:fix weight of auto-encoder, separately update weight of n classifiers(smaller learning rate)\\ 
3:update weight of all the model(smaller learning rate)\\
\begin{algorithm}[H]
\caption{Training scheme for MIRACLE}\label{alg:training-scheme}
\begin{algorithmic}[1]
\State Initialize the auto-encoder model and n multi-task classifiers\\
\Comment{\textbf{1st stage: Pretrain}:Iterate over each epoch.}
\For{$epoch$ in $num\_epochs\_1$} 
\State $Loss= \alpha \times MSE(\mathbf{x},D \circ E(\mathbf{x}))+\beta \times D_{KL}(E(\mathbf{x})) + \sum^t_{i=1}{\gamma_i\times BCE(C_{i} \circ E(\mathbf{x}),\mathbf{Y})}$
\State Update the weights of the auto-encoder and n multi-task classifiers.
\EndFor
\\
\Comment{\textbf{2nd stage: Finetune each task}:Iterate over each epoch.}
\For{$epoch$ in $num\_epochs\_2$} 
  \For{classifier in $n\_classifiers$} \Comment{Iterate over each multi-task classifier
    \State Fix the weight of the auto-encoder}
    \State $Loss= \sum^t_{i=1}{\gamma_i\times BCE(C_{i} \circ E(\mathbf{x}),\mathbf{Y})}$
    \State Update the weight of the current multi-task classifier using a smaller learning rate
  \EndFor
 \EndFor \\
 \Comment{\textbf{3rd stage: Finetune Whole}:Iterate over each epoch.}
 \For{$epoch$ in $num\_epochs\_3$} 
\State $Loss= \alpha \times MSE(\mathbf{x},D \circ E(\mathbf{x}))+\beta \times D_{KL}(E(\mathbf{x})) + \sum^t_{i=1}{\gamma_i\times BCE(C_{i} \circ E(\mathbf{x}),\mathbf{Y})}$
  \State Update the weights of all models using a smaller learning rate
 \EndFor 
\State \Return the trained model
\end{algorithmic}
\end{algorithm}
In the algorithm block above,
$dataset\_list$ means the list of training datasets, we have n datasets.
$num\_classifiers$ means the number of multi-task classifiers connected to the latent layer of auto-encoder.
$learning\_rate$ means the learning rate for updating all models in 1st stage.
$smaller\_learning\_rate$ means the smaller learning rate for updating classifiers in the latter stages.
$num\_epochs\_i$ means the number of epochs for training for $i-th$ stage.
 $E$ means encoder.$D$ means decoder. $C_i$means the $i-th$ classifier.

\subsubsection{Optimization: Multi-task weight assignment policy and learning rate scheduling}
We apply various optimization methods in multi-task learning, including MGDA\cite{Sener2018MultiTaskLA}, RW\cite{Lin2021ReasonableEO}, GroupDRO\cite{Sagawa2019DistributionallyRN}, and Gradnorm\cite{Chen2017GradNormGN}. Additionally, we introduce an algorithm, "PWInVal", that assigns weights to tasks inversely proportional to their validation accuracies, reducing multi-task learning loss.

PWInVal algorithm, detailed in the appendix, gives higher weights to tasks with validation accuracies below a threshold, prioritizing them during training. In contrast, tasks with accuracies at or above the threshold receive lower weights, enabling the model to focus more on underperforming tasks.

Our learning rate scheduling incorporates Reduce-LR-On-Plateau\cite{ChangeLROnPlateau}, which, depending on validation accuracy, reduces the learning rate by a factor when there's no improvement over certain epochs. This often results in enhanced performance, as a lower learning rate in the fine-tuning stage could better facilitate global optimum attainment.

% We use some optimization technique in the area of multi-task learning like MGDA(Multiple Gradient Descent Algorithm),RW(randomized weight). We reference the idea of GroupDRO,and Gradnorm. We also designed a multi-task weight assginment policy which inversely proportional to validation accuracy. \\
% We use some optimizater scheduling policy such as Reduce-LR-On-Plateau where the validation data accuracy is the metric, this scheduler will decrease learning rate by a factor when validation accuracy does not increase for some certain epoches. This scheduling policy sometimes outcomes better results. This may be explained by that in the finetune stage, the learning rate should be smaller to better reach the global optimal.

% In the following part, we present our multi-task learning task weight assignment algorithms "PWInVal","Piece-wise Inversely Proportional to Validation Accuracy". The detail of this algorithm can be found in appendix. The goal of the algorithm is to assign appropriate weights to each task, based on their validation accuracies, to facilitate lower loss during the multi-task learning.

% This policy ensures that tasks with validation accuracies below the threshold receive larger weights, making the model prioritize these tasks during training. Conversely, tasks that have reached or exceeded the threshold receive smaller weights, allowing the model to focus less on these tasks and more on tasks that have yet to reach the threshold.
\section{Experiments and Results}
\subsection{Experimental setup}
% \begin{table}[H]
% \small
% \begin{tabular}{|C|C|C|C|C|}
% \hline Disease & GSE ID & Samples & \makecell{\#positive case /\\ \# in test} & Notes \\

% \hline Type 1 diabetes & GSE142512 & 184 &23/53 & Final data: PBMC\\
% \hline \makecell{Inflammatory bowel\\ disease (IBD)} & GSE87648 & 379 &50/88 & \makecell{Ulcerative colitis and \\Crohn's disease labeled as +} \\
% \hline Multiple sclerosis (MS) & GSE106648 & 279 &34/79 & PBLs\\
% \hline Psoriasis & GSE73894 & 219 &35/49 & Skin tissues \\
% \hline Rheumatoid arthritis (RA) & GSE42861 & 689 &89/160 & peripheral blood leukocytes\\
% \hline \makecell{Systemic lupus \\erythematosus (SLE)} & GSE59250 & 343 & 
%  61/118 &\makecell{CD4+ T-cells, CD19+ B-cells \\and CD14+ Monocytes from\\ lupuspatients and controls.}\\
% \hline \makecell{Total} &    & 2093 & 281/547
%  &\makecell{}\\
% \hline
% \end{tabular}
% \end{table}
We employed six autoimmune disease-related methylome datasets for evaluation, as outlined in Table All datasets were profiled using the Illumina Infinium HumanMethylation450 BeadChip platform (platform ID GPL13534), which contains 482,421 CpG sites. The majority of the datasets were derived from blood samples, suggesting a possible commonality in gene or pathway level causation of the diseases. This characteristic makes it reasonable to adopt a multi-task learning architecture with a hard-shared common layer. Moreover, the format and dimension of the datasets are consistent, enabling the application of the site-gene-pathway explainable neural network.
The datasets encompass a range of autoimmune diseases as follows: Type 1 diabetes\cite{johnson2020longitudinal}, inflammatory bowel disease(IBD)\cite{ventham2016integrative}, multiple sclerosis(MS)\cite{kular2018dna} , psoriasis , rheumatoid arthritis (RA)\cite{liu2013epigenome}, systemic lupus erythematosus (SLE)\cite{absher2013genome}.
The details of dataset can be found in appendix "Dataset Details"

\subsection{Multi-task machine learning improves classification task accuracy}

In this section, we present the results of experiments conducted on six different datasets: Diabetes1, IBD, MS, Psoriasis, RA, and SLE. The table below displays the average accuracy and standard deviation (std) for various machine learning methods, including our proposed multi-task learning approach and baseline methods such as linear regression and logistic regression. The number of tests for each dataset is as follows: 53, 88, 79, 49, 160, 118.
% \usepackage{booktabs}
% \usepackage{booktabs}
% \usepackage{booktabs}% Please add the following required packages to your document preamble:
% \usepackage{booktabs}
% \usepackage{booktabs}
% \usepackage{booktabs}
% \usepackage{booktabs}

\begin{table}[H]
\centering
\scriptsize
\begin{tabular}{@{}ccccccccc@{}}
\toprule
\textbf{Method}                                                          & \begin{tabular}[c]{@{}c@{}}\textbf{Diabetes1}\\ \textbf{Accu(\%)}\end{tabular} & \begin{tabular}[c]{@{}c@{}}\textbf{IBD}\\ \textbf{Accu(\%)}\end{tabular}  & \begin{tabular}[c]{@{}c@{}}\textbf{MS}\\ \textbf{Accu(\%)}\end{tabular}    &\begin{tabular}[c]{@{}c@{}}\textbf{Psoriasis}\\ \textbf{Accu(\%)}\end{tabular}  & \begin{tabular}[c]{@{}c@{}}\textbf{RA}\\ \textbf{Accu(\%)}\end{tabular}  & \begin{tabular}[c]{@{}c@{}}\textbf{SLE}\\ \textbf{Accu(\%)}\end{tabular}  & \textbf{\begin{tabular}[c]{@{}c@{}}Average\\ Accuracy(\%)\end{tabular}} \\ \midrule

LinearRegression                                                         & 33.32$\pm$    /     & 30.27$\pm$    / & 13.22$\pm$    /  & 59.64$\pm$    /      & 31.94$\pm$    /  & 59.36$\pm$    /  & 37.50$\pm$    /  \\
KNeighbors                                                               & 84.78$\pm$10.84    & 75.79$\pm$8.11 & 65.71$\pm$7.32 & \textbf{90.91}$\pm$9.37 & 74.57$\pm$5.69 & 87.16$\pm$4.69 & 78.65$\pm$6.93 \\
LogisticRegression                                                       & 86.96$\pm$7.73     & 74.74$\pm$6.37 & 68.57$\pm$8.78 & 89.09$\pm$10.65    & 76.88$\pm$4.64 & 88.99$\pm$8.07 & 80.02$\pm$7.09 \\
BernoulliNB                                                              & 65.22$\pm$12.27    & 46.32$\pm$1.08 & 48.57$\pm$1.77 & 76.36$\pm$2.32     & 54.91$\pm$0.69 & 50.46$\pm$0.91 & 54.57$\pm$2.23 \\
GaussianNB                                                               & 82.61$\pm$4.45     & 77.89$\pm$7.40 & 68.57$\pm$9.05 & 83.64$\pm$8.36     & 75.72$\pm$4.58 & 88.99$\pm$6.65 & 79.28$\pm$6.45 \\
DecisionTree                                                             & 76.09$\pm$14.67    & 65.26$\pm$9.75 & 55.71$\pm$11.25 & 81.82$\pm$11.77    & 68.79$\pm$6.42 & 78.90$\pm$7.44 & 70.39$\pm$9.15 \\
SVC                                                                      & 91.30$\pm$8.89     & 76.84$\pm$5.93 & 70.00$\pm$10.48 & 87.27$\pm$11.42    & 77.46$\pm$5.64 & 88.99$\pm$6.38 & 80.99$\pm$7.38 \\
AdaBoost                                                                 & 82.61$\pm$5.47     & 62.11$\pm$10.12 & 64.29$\pm$10.69 & 83.64$\pm$8.95 & 73.41$\pm$6.83 & 84.40$\pm$7.79 & 74.45$\pm$8.18 \\
GradientBoosting & 91.30$\pm$8.35 & 69.47$\pm$11.33 & 62.86$\pm$9.56 & 87.27$\pm$9.58 & 74.57$\pm$6.22 & 85.32$\pm$7.28 & 77.14$\pm$8.26 \\
RandomForest & 89.13$\pm$5.53 & 74.74$\pm$5.85 & 71.43$\pm$11.24 & 87.27$\pm$10.03 & 77.46$\pm$6.27 & 86.24$\pm$6.87 & 80.05$\pm$7.32 \\
\begin{tabular}[c]{@{}c@{}}MulticlassNN\\ (Top25 residue)\end{tabular} & 84.91$\pm$8.42 & 72.73$\pm$7.23 & 74.68$\pm$9.10 & 71.43$\pm$9.13 & 73.75$\pm$5.16 & 87.29$\pm$6.42 & 77.51$\pm$7.01 \\
\begin{tabular}[c]{@{}c@{}}MulticlassNN\\ (Top1000 residue)\end{tabular} & \textbf{96.23}$\pm$8.51 & 67.05$\pm$7.34 & 60.76$\pm$9.14 & 71.43$\pm$8.95 & 73.13$\pm$5.23 & 42.37$\pm$6.22 & 65.81$\pm$7.00 \\
\begin{tabular}[c]{@{}c@{}}MulticlassNN\\ (Top2000 residue)\end{tabular} & 67.93$\pm$8.04 & 56.82$\pm$8.12 & 56.96$\pm$10.07 & 71.43$\pm$9.77 & 68.75$\pm$5.79 & 51.70$\pm$6.88 & 61.61$\pm$7.59 \\
\begin{tabular}[c]{@{}c@{}}\textbf{Multi-task}\\ \textbf{(pretrain)(Ours)}\end{tabular} & 92.00$\pm$8.04 & 73.83$\pm$6.88 & 77.59$\pm$9.31 & 75.38$\pm$9.48 & 85.53$\pm$5.67 & 96.52$\pm$6.76 & 84.10$\pm$7.32 \\
\begin{tabular}[c]{@{}c@{}}\textbf{Multi-task}\\ \textbf{(finetune)(Ours)}\end{tabular} & 88.00$\pm$8.20 & \textbf{79.44}$\pm$7.14 & \textbf{77.59}$\pm$9.43 & 89.23$\pm$9.53 & \textbf{86.18}$\pm$5.68& \textbf{99.13}$\pm$6.81 & \textbf{87.20}$\pm$7.29 \\ \bottomrule
\end{tabular}
%\end{tabularx}
\end{table}

Our proposed multi-task learning method with fine-tuning achieved the highest average accuracy (87.20\%) across all datasets, with a standard deviation of 7.29. It outperformed all other methods in the IBD, MS, RA, and SLE datasets. However, our method did not achieve the highest accuracy in the Diabetes1 and Psoriasis datasets. A possible explanation for this result is that the Diabetes1 and Psoriasis datasets have the smallest number of samples among all datasets. In multi-task learning, having fewer samples in a task may affect the model's ability to learn shared representations effectively, leading to a decrease in performance. To address this issue in future work, we can consider increasing the task weight for these smaller datasets, which can help the model focus more on these tasks during training. Alternatively, we can explore other multi-task learning techniques that have been proposed in the literature to improve performance in cases with imbalanced sample sizes.

In summary, our proposed multi-task learning approach demonstrates superior performance in almost all datasets compared to other methods, with some room for improvement in the Diabetes1 and Psoriasis datasets. These results highlight the effectiveness of our method for handling a wide range of tasks and the potential for further enhancements.
\subsection{MIRACLE can detect common embedding features between multiple diseases}

\begin{figure}[H]
\centering
\includegraphics[width=1.1\linewidth]{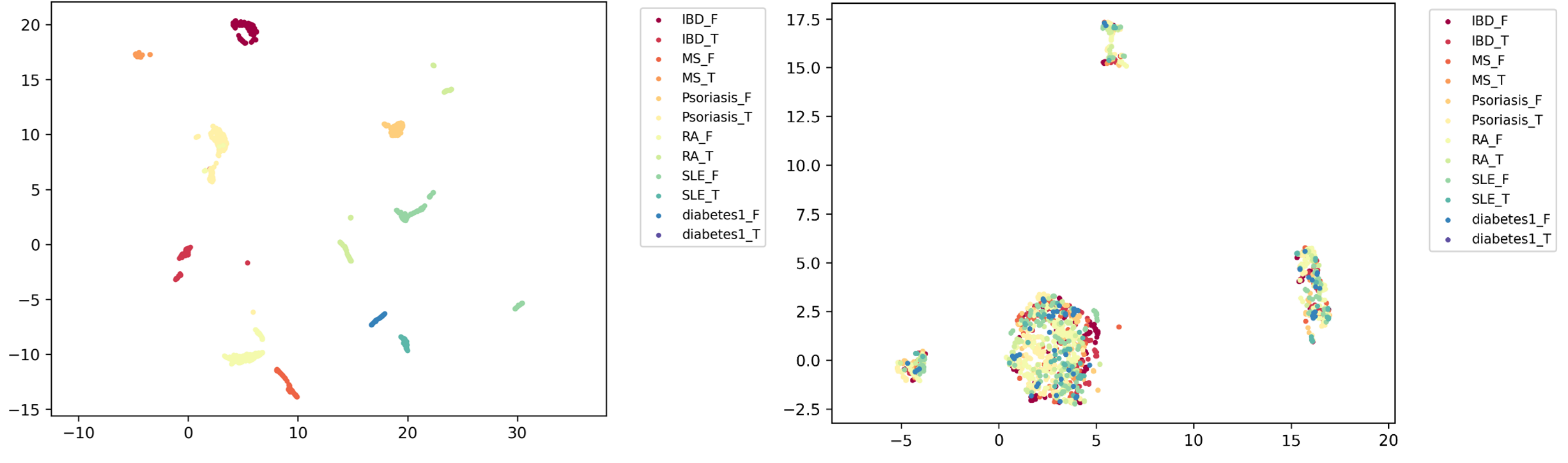}
    \caption{Left: UMAP projection of raw features for all samples in six datasets.  Right: UMAP projection of embedding features by MIRACLE(epoch 999 of second training stage, number of site=200)}
\centering
\end{figure}
Exploring commonalities among multiple autoimmune diseases is crucial. Generative models provide us with a perspective on the generative paradigm between multiple datasets. In this study, we use VAE as a tool to explore commonalities among multiple autoimmune diseases by jointly training multiple datasets to prevent overfitting. The embedding layer represents common feature patterns among multiple disease datasets, which are both more consistent with biological priors and provide a unified research perspective for exploring autoimmune diseases.

We compared the similarity of samples under unsupervised conditions between original features and MIRACLE’s embedding layer and visualized them using UMAP. We found that under original features, samples showed the characteristics of separation into multiple different types of samples, and there was no correlation between multiple categories and positive/negative samples. Therefore, under original features, we focus more on the degree of feature differentiation between categories, and it is difficult to observe the correlation between multiple disease samples. Through MIRACLE training, the embedding layer clustered samples into four clusters, so we performed feature engineering on multiple different types of samples and mapped them to the feature space expressing commonality of autoimmune diseases. Moreover, benchmark comparisons also proved that the commonality feature space has improved subsequent classification. Therefore, it proves the importance of exploring common mechanisms among multiple diseases.

\section{Conclusions}
In this paper, we propose a multi-task learning-based prediction classification model for DNA methylation in autoimmune diseases based on epigenetics. The model analyzes the commonalities between autoimmune diseases by jointly training multiple datasets to solve the problem of few samples and many features in a single dataset. We use the VAE model as a common expression generator model for autoimmune diseases and design a three-layer (residue-gene-pathway) generative regulatory network. We add regularization terms to the connections between neurons to make the network interpretable according to biological prior knowledge (Go ontology). Promising results demonstrate the feasibility of our method in single disease prediction and multi-disease commonality research, providing new perspectives for the mechanism research of epigenomics and disease relationships.

\newpage

% References follow the acknowledgments in the camera-ready paper. Use unnumbered first-level heading for
% the references. Any choice of citation style is acceptable as long as you are
% consistent. It is permissible to reduce the font size to \verb+small+ (9 point)
% when listing the references.
% Note that the Reference section does not count towards the page limit.
\medskip

\bibliographystyle{unsrt}
\bibliography{ref}

\newpage
\appendix

\section{motivation}
\begin{figure}[htbp]

    \begin{minipage}[b]{1\linewidth}
        \centering
        \includegraphics[width=0.9\linewidth]{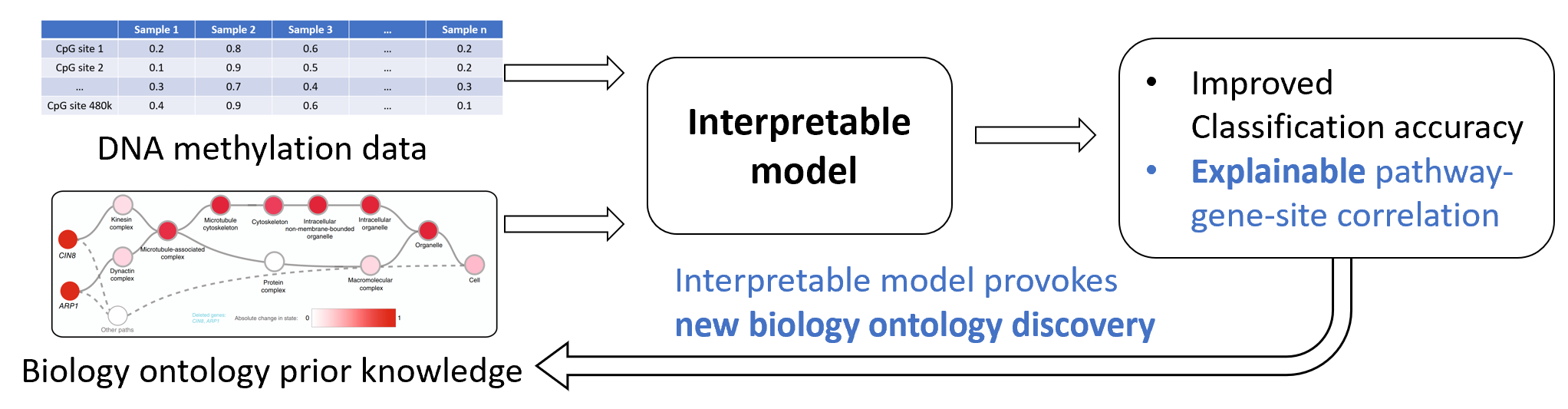}
   \caption{Our goal}
    \end{minipage}

\end{figure}
Despite significant advances in the field of DNA methylation research, there remains considerable ambiguity in understanding the role of DNA methylation in the development and progression of autoimmune diseases. One of the primary challenges faced by researchers is the limited availability of sample data for these diseases. Consequently, there is an urgent need for a robust method capable of integrating and analyzing data across multiple sources to provide a more comprehensive understanding of the underlying biological mechanisms.

The primary goal of this study is to develop an interpretable model based on prior knowledge that can accurately classify DNA methylation data and facilitate the discovery of common features across different autoimmune diseases. To achieve this objective, we propose MIRACLE, a novel interpretable neural network that leverages autoencoder-based multi-task learning to effectively connect and analyze multiple datasets.
The goal of this study is to build an interpretable model based on prior knowledge to classify DNA methylation data. To achieve this goal, a novel interpretable neural network, MIRACLE, is proposed. MIRACLE is an auto-encoder-based multi-task learning neural network that connects multiple datasets and is trained together to identify common features in DNA methylation.

%\section{Appendix}
%The source code of this project is at 
%  \url{https://github.com/explcre/Adaptable-and-intrepretable-multi-task-learning-based-gene-level-methylation-estimation}
\newpage
\section{Code of MaskedLinear Layer}
Here is a Python code to explain how is this layer implemented:
\begin{lstlisting}[language=Python, caption=MaskedLinear Class Definition example]
class MaskedLinear(nn.Linear):
    def __init__(self, *args, mask, **kwargs):
        super().__init__(*args, **kwargs)
        self.mask = mask
    def forward(self, input):
        self.masked_weight = self.weight * self.mask    
        return torch.nn.functional.linear(input, self.masked_weight, bias=self.bias)
\end{lstlisting}
\begin{lstlisting}[language=Python, caption=MaskedLinear Calling example]
self.encoder = nn.Sequential(
    MaskedLinear(gene_layer_dim, latent_dim, mask=self.pathway_gene_tensor.T),
    nn.Sigmoid()
)
\end{lstlisting}
\section{Convergence of MaskedLinear Layer}

To analyze the convergence of this method, we can use the framework and results from the "On the Convergence of Shallow Neural Network Training with Randomly Masked Neurons" paper. Specifically, we can apply the concepts of training with randomly masked neurons, surrogate functions, and convergence analysis to the MaskedLinear method.

The MaskedLinear method can be interpreted as a single hidden-layer neural network with ReLU activations and masked weights:

$$
f(\mathbf{W}, \mathbf{a}, \mathbf{x})=\frac{1}{\sqrt{m}} \sum_{r=1}^m a_r \sigma\left(\left\langle\mathbf{w}_r, \mathbf{x}\right\rangle\right)
$$
The masked weights can be represented as:
$$
f_{\mathbf{m}_k^l}(\mathbf{W}, \mathbf{x})=\frac{1}{\sqrt{m}} \sum_{r=1}^m a_r m_{k, r}^l \sigma\left(\left\langle\mathbf{w}_r, \mathbf{x}\right\rangle\right)
$$
From the \cite{DBLP:journals/corr/abs-2112-02668}"On the Convergence of Shallow Neural Network Training with Randomly Masked Neurons" paper, we know that the convergence behavior is controlled by the NTK matrix:
$$
\mathbf{H}(k)_{i j}=\frac{\xi}{m}\left\langle\mathbf{x}_i, \mathbf{x}_j\right\rangle \sum_{r=1}^m \mathbb{I}\left\{\left\langle\mathbf{w}_{k, r}, \mathbf{x}_i\right\rangle \geq 0,\left\langle\mathbf{w}_{k, r}, \mathbf{x}_j\right\rangle \geq 0\right\}
$$
The convergence of the MaskedLinear method can thus be analyzed by considering the NTK matrix and how it evolves during training. In the overparameterized regime, the change of the network's weights is controlled in a small region around initialization, and the change of $\mathbf{H}(k)$ is small, staying close to the NTK at initialization. This allows the network to converge to a good solution.

The representational ability of the MaskedLinear method can also be analyzed using the framework from the "On the Convergence of Shallow Neural Network Training with Randomly Masked Neurons" paper. The paper shows that the convergence of the network is related to the structure of the dataset and the activation function. In the case of the MaskedLinear method, the activation function is a linear function, which simplifies the analysis. The representational ability of the network will depend on the ability to capture the underlying structure of the input site features $X_{site}$ and output pathway features $Y_{pathway}$.

In summary, the convergence and representational ability of the MaskedLinear method can be better than original Linear layer.
\newpage
\section{Data Pre-processing Algorithm}
If we use all of 480k neuron to build the neural network, the size of the neural network will be extremely large. Therefore, for the purpose of decrease the complexity of the model and add computation, we use some data pre-processing algorithms to filter out unimportant features.

\begin{algorithm}
\caption{Selection of significant sites from a list of datasets}\label{alg:significant-sites}
\begin{algorithmic}[1]
\Require $dataset\_list$: a list of datasets
\Require $num\_selected\_site$: the number of significant sites to select (optional)
\Ensure $site\_set$: a set of significant sites
\State $site\_set$ $\gets$ $\{\}$ \Comment{Initialize the set of significant sites}
\For{$dataset\_i$ in $dataset\_list$} \Comment{Iterate over each dataset}
\State Perform a t-test of positive and negative samples in $dataset\_i$
\State Sort the p-values ascendingly
\State Drop the p-value which is larger than 0.05, or select $num\_selected\_site$ smallest p-values
\State Add the remained site names into the $site\_set$
\EndFor
\State \Return $site\_set$
\end{algorithmic}
\end{algorithm}
\section{Multi-task Learning Task Weight Assignment Policy}
\subsubsubsection{\textbf{Piece-wise Inversely Proportional to Validation Accuracy Multi-task Weight Assignment}}

Formally, let $w_i$ be the weight assigned to task $i$, $\text{val\_acc}_i$ be the validation accuracy of task $i$, $s_i$ be the single-task accuracy upper bound for task $i$, and $W$ be the \textit{SINGLE\_TASK\_UPPER\_BOUND\_WEIGHT}. The weight assignment function can be expressed as:
\begin{algorithm}[H]
\caption{Piece-wise Inversely Proportional to Validation Accuracy Multi-task Weight Assignment}
\begin{algorithmic}[1]
\State Initialize model parameters $\Theta$
\For{each epoch}
    \For{each mini-batch}
        \For{each task $i$ in tasks}
            \State Compute the loss $L_i$ for task $i$
            \State Compute the validation accuracy $\text{val\_acc}_i$ for task $i$
        \EndFor
        \For{each task $i$ in tasks}
            \If{$\text{val\_acc}_i \leq s_i$}
                \State $w_i \gets \frac{(W - 1)}{s_i} \cdot \text{val\_acc}_i + 1$
            \Else
                \State $w_i \gets -\frac{W}{(s_i - 1)} \cdot \text{val\_acc}_i + \frac{W}{(1 - s_i)}$
            \EndIf
        \EndFor
        \State Update model parameters $\Theta$ using combined loss $\sum_{i} w_i L_i$
    \EndFor
\EndFor
\end{algorithmic}
\end{algorithm}

This policy ensures that tasks with validation accuracies below the threshold receive larger weights, making the model prioritize these tasks during training. Conversely, tasks that have reached or exceeded the threshold receive smaller weights, allowing the model to focus less on these tasks and more on tasks that have yet to reach the threshold.
\newpage 
\section{Dataset Details}
\begin{table}[H]
\small
\begin{tabular}{|c|c|c|c|c|}
\hline Disease & GSE ID & Samples & \makecell{\#positive case /\\ \# in test} & Notes \\

\hline Type 1 diabetes & GSE142512 & 184 &23/53 & Final data: PBMC\\
\hline \makecell{Inflammatory bowel\\ disease (IBD)} & GSE87648 & 379 &50/88 & \makecell{Ulcerative colitis and \\Crohn's disease labeled as +} \\
\hline Multiple sclerosis (MS) & GSE106648 & 279 &34/79 & PBLs\\
\hline Psoriasis & GSE73894 & 219 &35/49 & Skin tissues \\
\hline Rheumatoid arthritis (RA) & GSE42861 & 689 &89/160 & peripheral blood leukocytes\\
\hline \makecell{Systemic lupus \\erythematosus (SLE)} & GSE59250 & 343 & 
 61/118 &\makecell{CD4+ T-cells, CD19+ B-cells \\and CD14+ Monocytes from\\ lupuspatients and controls.}\\
\hline \makecell{Total} &    & 2093 & 281/547
 &\makecell{}\\
\hline
\end{tabular}
\end{table}

The datasets encompass a range of autoimmune diseases as follows: Type 1 diabetes, inflammatory bowel disease(IBD), multiple sclerosis(MS) , psoriasis , rheumatoid arthritis (RA), systemic lupus erythematosus (SLE).

As follows:

\textbf{GSE142512:} Genome-wide DNA methylation profiling of peripheral blood mononuclear cells (PBMC) from 184 Type 1 diabetes patients and controls [36].

\textbf{GSE87648:} DNA methylation analysis of 379 inflammatory bowel disease (IBD) patients, including ulcerative colitis and Crohn's disease, and healthy controls [37].

\textbf{GSE106648:} Genome-wide DNA methylation profiling of peripheral blood leukocytes (PBLs) from 279 multiple sclerosis (MS) patients and controls [38].

\textbf{GSE73894:} DNA methylation profiling of skin tissues from 219 psoriasis patients and healthy controls [39].

\textbf{GSE42861:} Genome-wide DNA methylation profiling of peripheral blood leukocytes from 689 rheumatoid arthritis (RA) patients and controls [46, 48].

\textbf{GSE59250:} Genome-wide DNA methylation analysis of CD4+ T-cells, CD19+ B-cells, and CD14+ monocytes from 343 systemic lupus erythematosus (SLE) patients and controls [40].

These datasets were selected for their relevance to autoimmune diseases and their compatibility with our multi-task learning architecture.
We use the datasets as above. They are all Autoimmune disease. They are from the Infinium Human Methylation 450k BeadChip data,482,421 CpG sites. And most of the datasets is tested by blood. Therefore, we can assume that they have some common cause of the disease in the gene level or pathway level. This characteristic of datasets make it reasonable to choose the architecture of multi-task learning with hard share common layer. 
Also, the format and dimension of the dataset is the same, make the site-gene-pathway explainable neural network workable.

\newpage

\section{Supplementary Results}
\subsection{Plot of 6 dataset prediction accuracy results}
\begin{figure}[H]
        \centering
        \includegraphics[width=1\linewidth]{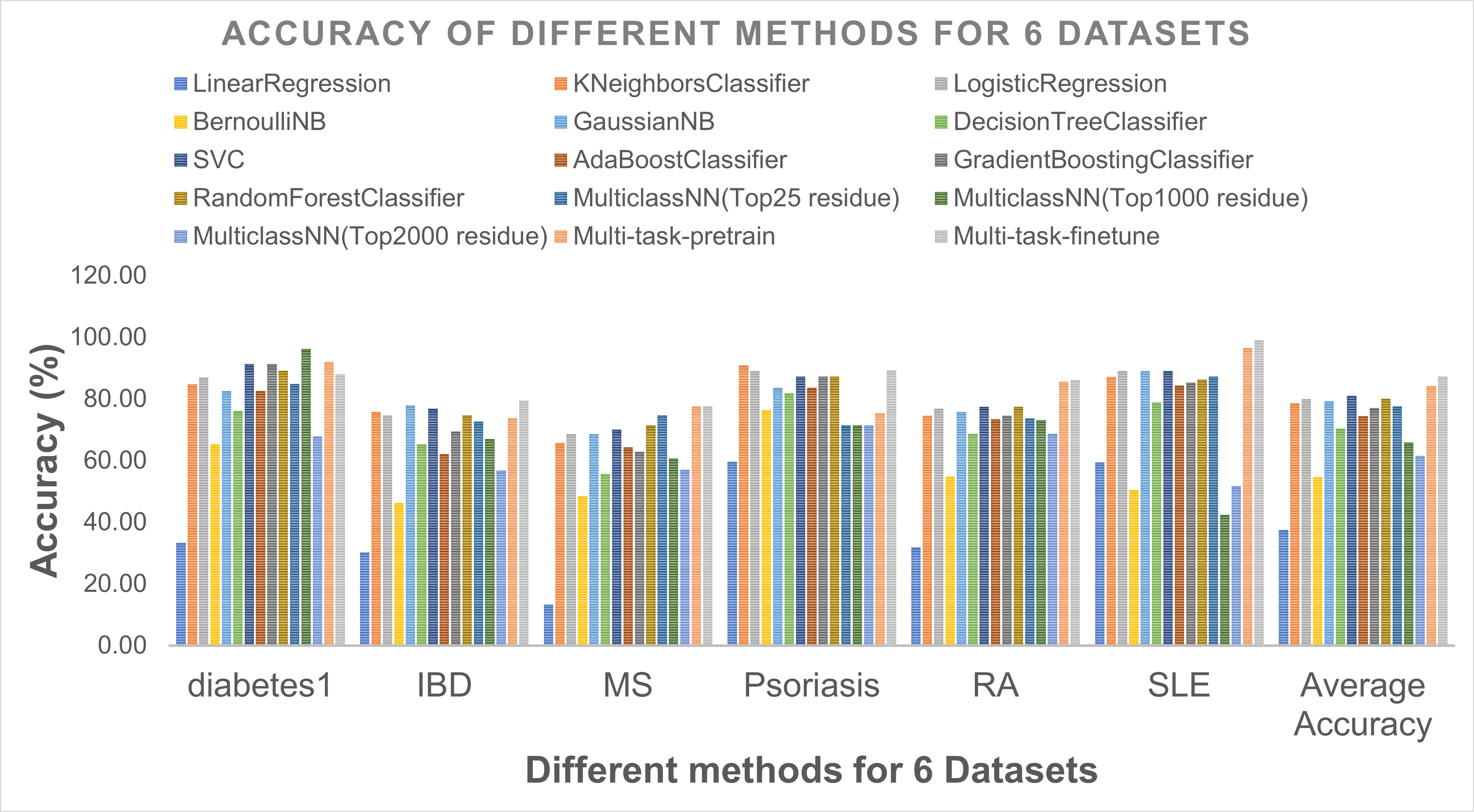}
   \caption{Accuracy of different algorithms for 6 datasets}
\end{figure}
\subsection{Abalation Study}
% \usepackage{booktabs}
\begin{table}[H]
\scriptsize
\centering
\begin{tabular}{@{}cccccccc@{}}
\toprule
model-type                                                   & diabetes1      & IBD            & MS             & Psoriasis      & RA             & SLE            & total\_accuracy \\ \midrule
Multi-task VAE full-connected NN with dropout\_ratio=0.5     & 90.00          & 67.29          & 74.14          & 90.77          & 84.87          & 95.65          & 83.73           \\
Single Task VAE site-gene-pathway explainable NN(finetuned)  & 90.00          & 73.83          & 75.86          & 81.54          & \textbf{86.84} & 98.26          & 85.19           \\
Multi-task VAE site-gene-pathway explainable NN (pretrained) & \textbf{92.00} & 73.83          & 77.59          & 75.38          & 85.53          & 96.52          & 84.10           \\
Multi-task VAE site-gene-pathway explainable NN (finetuned)  & 88.00          & \textbf{97.44} & \textbf{77.59} & \textbf{89.23} & 86.18          & \textbf{99.13} & \textbf{87.20}  \\ \bottomrule
\end{tabular}
\end{table}
\subsubsection{Effectiveness of MaskedLinear site-gene-pathway explainable layer compared to Random Dropout}
We compared our work with exactly the samme architecture but the explainable maskedLinear layer is replaced by fully-connected layer with random dropout.
Our explainable site-gene-pathway layer can have better accuracy than simply randomly dropout the fully-connected layer. This is because our site-gene-pathway uses the prior knowledge of gene ontology, which is already discovered by biology research. Thus, the explainable structure can not only have better accuracy, but also give us a human understandable model if we check certain weight of certain neurons connections between site-gene or gene-pathway layers.
\subsubsection{Effectiveness of Multi-task Learning compared to Single Task}
We compared our work with exactly the samme architecture but the explainable maskedLinear layer is replaced by fully-connected layer with random dropout.
We use six auto-immune diseases to fit a multi-task learning neural network. We can see from the table that our method can works better than most of the single-task laerning with exactly same structure but only one downstream classifier. This is because the multi-task learning can facilite the mutual information between six diseases and better predict the common pattern between these six auto-immune diseases.
\newpage
\subsection{Interpretable MIRACLE's Pathway-Gene-Site Feature Relationship Neural Network}

In this section, we present a visualization of the MaskedLinear layers and examine the weight distribution of these layers across different training epochs. The MaskedLinear layer is a linear layer modified by the adjacency matrices of the site-gene and gene-pathway relationships, ensuring the preservation of gene ontology explainability.

We extract weights from the site-gene or gene-pathway MaskedLinear layers. In this discussion, we primarily focus on the results of the site-gene layer, although the gene-pathway layer exhibits a similar pattern. First, we define the terms masked distribution, ones distribution, and non-ones distribution. The ones distribution refers to positions where the site-gene/gene-pathway connection is 1, indicating a connection. We mask a certain percentage of the ground truth for these positions, treating them as unknown to assess whether the model can discover new connections. The distribution of these positions in the weight matrix constitutes the masked distribution. All other positions are part of the non-ones distribution.

As the number of epochs increases, the weight distribution of the Site-Gene and Gene-Pathway MaskedLinear layers exhibits a trimodal distribution. Upon examining the connections using known biological research, we find that they correspond to already discovered strong connections. Furthermore, when we mask portions of the site-gene-pathway information, MIRACLE demonstrates its ability to predict new biomarkers based on the MaskedLinear weight information.

\begin{figure}[H]
\subfigure
{
    \begin{minipage}[b]{0.3\linewidth}
        \centering
        \includegraphics[width=1\linewidth]{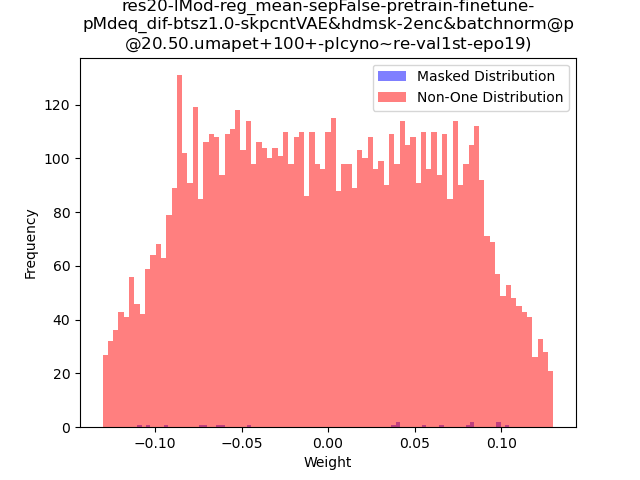}
   \caption{site=20 epoch=19 site-gene layer masked and non-one distribution}
    \end{minipage}
}
\subfigure
{
    \begin{minipage}[b]{0.3\linewidth}
        \centering
        \includegraphics[width=1\linewidth]{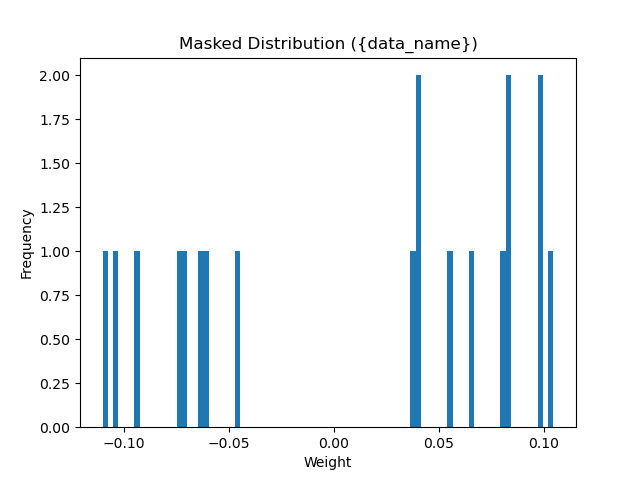}
   \caption{site=20 epoch=19 site-gene layer masked distribution}
    \end{minipage}
}
\subfigure
{
    \begin{minipage}[b]{0.3\linewidth}
        \centering
        \includegraphics[width=1\linewidth]{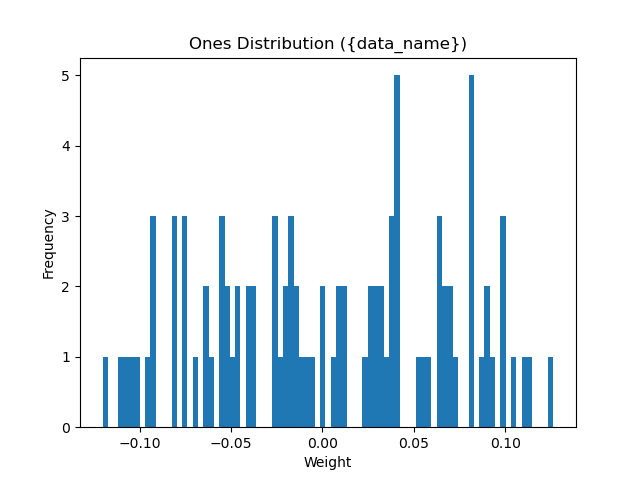}
   \caption{site=20 epoch=19 site-gene layer ones distribution}
    \end{minipage}
}
\end{figure}
\begin{figure}[H]
\subfigure
{
    \begin{minipage}[b]{0.3\linewidth}
        \centering
        \includegraphics[width=1\linewidth]{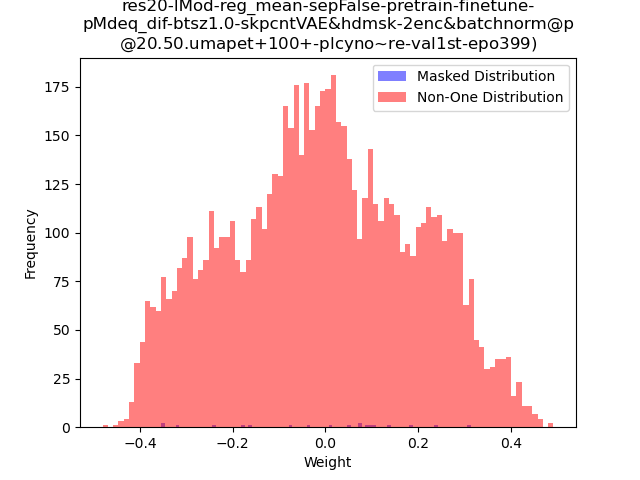}
   \caption{site=20 epoch=399 site-gene layer masked and non-one distribution}
    \end{minipage}
}
\subfigure
{
    \begin{minipage}[b]{0.3\linewidth}
        \centering
        \includegraphics[width=1\linewidth]{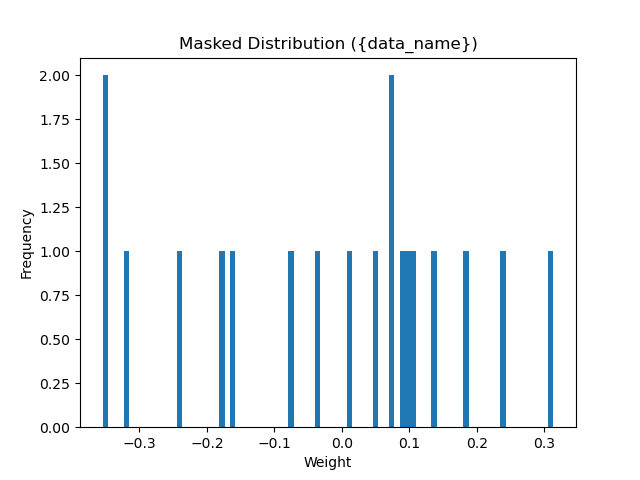}
   \caption{site=20 epoch=399 site-gene layer masked  distribution}
    \end{minipage}
}
\subfigure
{
    \begin{minipage}[b]{0.3\linewidth}
        \centering
        \includegraphics[width=1\linewidth]{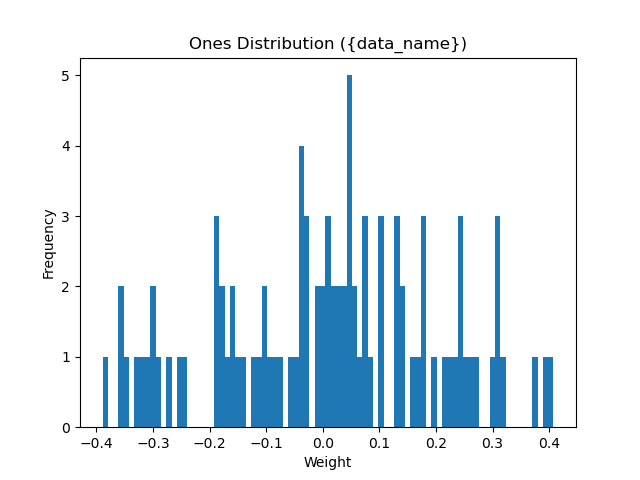}
   \caption{site=20 epoch=399 site-gene layer ones distribution}
    \end{minipage}
}
\end{figure}
\begin{figure}[H]
\subfigure
{
    \begin{minipage}[b]{0.3\linewidth}
        \centering
        \includegraphics[width=1\linewidth]{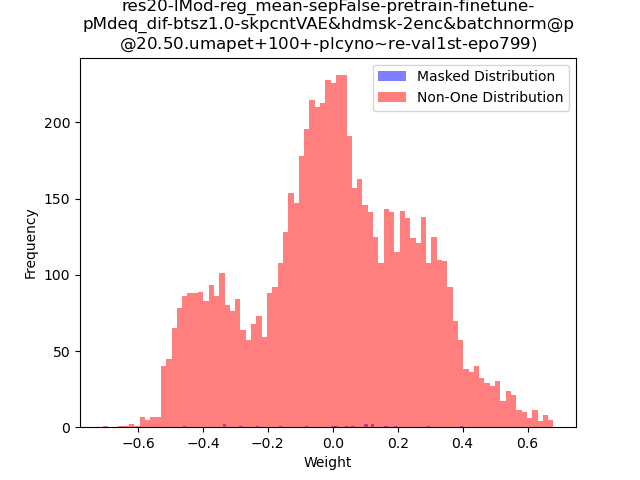}
   \caption{site=20 epoch=799 site-gene layer masked and non-one distribution}
    \end{minipage}
}
\subfigure
{
    \begin{minipage}[b]{0.3\linewidth}
        \centering
        \includegraphics[width=1\linewidth]{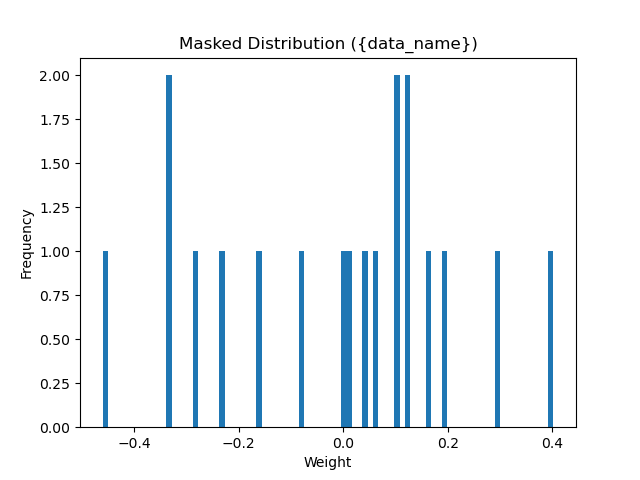}
   \caption{site=20 epoch=799 site-gene layer masked distribution}
    \end{minipage}
}
\subfigure
{
    \begin{minipage}[b]{0.3\linewidth}
        \centering
        \includegraphics[width=1\linewidth]{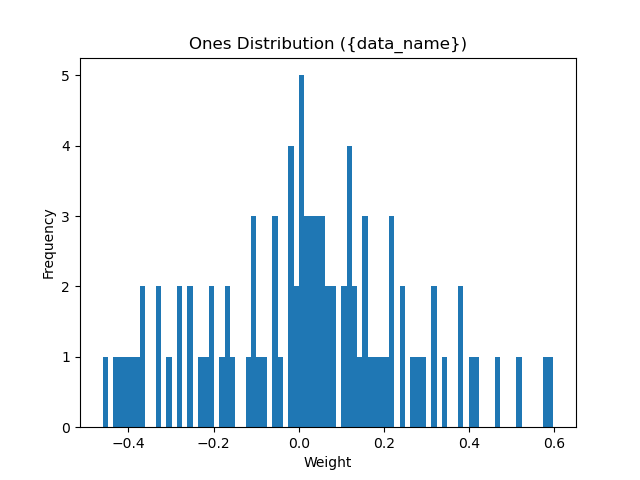}
   \caption{site=20 epoch=799 site-gene layer ones distribution}
    \end{minipage}
}
\end{figure}
For certain diseases, MIRACLE has identified novel relationships between pathways and genes, suggesting that some genes may have a significant association with the expression of specific diseases.
\subsection{MIRACLE discovered the correspondence between new pathways and genes}

For some diseases, MIRACLE has discovered the relationship between new pathways and genes, proving that some genes may have an important relationship with the expression of certain diseases

\newpage
\subsection{MIRACLE other neural network architecture results}
\begin{table}[H]
\tiny
\tabcolsep=0.03cm
\begin{tabular}{lllllllll
>{\columncolor[HTML]{FFFF00}}l llllll
>{\columncolor[HTML]{FFFF00}}l llllll
>{\columncolor[HTML]{FFFF00}}l ll}
\makecell{$\#site$} & $\#epo$ & model & diab1 & IBD     & MS      & Ps & RA       & SLE     & \makecell{total\\$accu$} & $dia1_2$ & $IBD_2$ & $MS_2$ & $Ps_2$ & $RA_2$ & $SLE_2$ & \makecell{total\\$accu_2$}& $dia1_3$ & $IBD_3$ & $MS_4$ & $Ps_3$ & $RA_3$ & $SLE_3$ & \makecell{total\\$accu_3$} & \multicolumn{2}{l}{policy} \\
20              & 125     & unet       & 69.32\%   & 77.22\% & 71.43\% & 73.13\%   & 84.75\%  & 84.91\% & 76.60\%         & 71.59\%           & 77.22\%     & 79.59\%    & 73.13\%           & 83.90\%    & 83.02\%     & 77.33\%                 & 65.91\%          & 74.68\%    & 79.59\%   & 73.75\%          & 84.75\%   & 83.02\%    & 76.42\%                & no                       &                        \\
20              & 200     & unet       & 81.82\%   & 67.09\% & 65.31\% & 76.25\%   & 82.20\%  & 81.13\% & 76.60\%         & 76.14\%           & 70.89\%     & 65.31\%    & 75.00\%           & 83.05\%    & 83.02\%     & 76.23\%                 & 76.14\%          & 70.89\%    & 65.31\%   & 74.38\%          & 83.90\%   & 81.13\%    & 76.05\%                & no                       &                        \\
20              & 200     & no         & 70.45\%   & 65.82\% & 71.43\% & 76.88\%   & 86.44\%  & 69.81\% & 75.14\%         & 71.59\%           & 69.62\%     & 71.43\%    & 77.50\%           & 85.59\%    & 69.81\%     & 75.87\%                 & 70.45\%          & 67.09\%    & 71.43\%   & 77.50\%          & 86.44\%   & 69.81\%    & 75.50\%                & no                       &                        \\
20              & 300     & unet       & 76.14\%   & 65.82\% & 79.59\% & 79.38\%   & 83.90\%  & 77.36\% & 77.70\%         & 70.45\%           & 67.09\%     & 79.59\%    & 79.38\%           & 83.05\%    & 79.25\%     & 76.97\%                 & 76.14\%          & 64.56\%    & 81.63\%   & 79.38\%          & 83.90\%   & 79.25\%    & 77.88\%                & no                       &                        \\
20              & 300     & no         & 71.59\%   & 65.82\% & 77.55\% & 78.75\%   & 84.75\%  & 67.92\% & 75.87\%         & 72.73\%           & 70.89\%     & 75.51\%    & 78.13\%           & 84.75\%    & 69.81\%     & 76.60\%                 & 71.59\%          & 65.82\%    & 75.51\%   & 78.75\%          & 83.90\%   & 69.81\%    & 75.69\%                & no                       &                        \\
100             & 300     & unet       & 73.86\%   & 50.63\% & 73.47\% & 80.00\%   & 91.53\%  & 86.79\% & 77.33\%         & 71.59\%           & 82.28\%     & 79.59\%    & 78.75\%           & 88.14\%    & 88.68\%     & 81.17\%                 & 75.00\%          & 70.89\%    & 75.51\%   & 79.38\%          & 91.53\%   & 88.68\%    & 80.62\%                & no                       &                        \\
100             & 300     & no         & 76.14\%   & 65.82\% & 75.51\% & 80.63\%   & 89.83\%  & 92.45\% & 80.44\%         & 71.59\%           & 81.01\%     & 77.55\%    & 80.63\%           & 92.37\%    & 86.79\%     & 82.08\%                 & 76.14\%          & 70.89\%    & 73.47\%   & 79.38\%          & 90.68\%   & 90.57\%    & 80.62\%                & no                       &                        \\
20              & 500     & unet       & 72.73\%   & 67.09\% & 81.63\% & 80.00\%   & 84.75\%  & 84.91\% & 78.61\%         & 75.00\%           & 67.09\%     & 83.67\%    & 82.50\%           & 83.05\%    & 83.02\%     & 79.34\%                 & 75.00\%          & 67.09\%    & 81.63\%   & 82.50\%          & 84.75\%   & 81.13\%    & 79.34\%                & no                       &                        \\
20              & 500     & no         & 78.41\%   & 72.15\% & 71.43\% & 80.00\%   & 84.75\%  & 83.02\% & 79.16\%         & 77.27\%           & 73.42\%     & 71.43\%    & 78.75\%           & 83.90\%    & 83.02\%     & 78.61\%                 & 78.41\%          & 72.15\%    & 71.43\%   & 80.00\%          & 86.44\%   & 83.02\%    & 79.52\%                & no                       &                        \\
20              & 1000    & unet       & 78.41\%   & 69.62\% & 79.59\% & 81.25\%   & 84.75\%  & 73.58\% & 78.98\%         & 79.55\%           & 73.42\%     & 85.71\%    & 80.00\%           & 84.75\%    & 71.70\%     & 79.71\%                 & 79.55\%          & 70.89\%    & 79.59\%   & 81.25\%          & 84.75\%   & 71.70\%    & 79.16\%                & no                       &                        \\
20              & 1000    & no         & 77.27\%   & 77.22\% & 81.63\% & 81.25\%   & 84.75\%  & 88.68\% & 81.54\%         & 68.18\%           & 74.68\%     & 79.59\%    & 81.88\%           & 84.75\%    & 84.91\%     & 79.34\%                 & 67.05\%          & 73.42\%    & 79.59\%   & 82.50\%          & 84.75\%   & 83.02\%    & 78.98\%                & no                       &                        \\
100             & 1000    & unet       & 69.32\%   & 75.95\% & 77.55\% & 83.75\%   & 93.22\%  & 88.68\% & 82.27\%         & 67.05\%           & 74.68\%     & 77.55\%    & 83.75\%           & 93.22\%    & 88.68\%     & 81.72\%                 & 70.45\%          & 77.22\%    & 79.59\%   & 83.13\%          & 91.53\%   & 88.68\%    & 82.27\%                & no                       &                        \\
100             & 1000    & no         & 65.91\%   & 77.22\% & 77.55\% & 85.63\%   & 93.22\%  & 90.57\% & 82.63\%         & 68.18\%           & 75.95\%     & 79.59\%    & 82.50\%           & 94.07\%    & 90.57\%     & 82.27\%                 & 68.18\%          & 75.95\%    & 77.55\%   & 83.75\%          & 93.22\%   & 90.57\%    & 82.27\%                & no                       &                        \\
2000            & 500     & unet       & 60.23\%   & 54.43\% & 75.51\% & 71.25\%   & 82.20\%  & 56.60\% & 68.37\%         & 70.45\%           & 69.62\%     & 71.43\%    & 80.00\%           & 90.68\%    & 56.60\%     & 76.23\%                 & 67.05\%          & 72.15\%    & 75.51\%   & 78.13\%          & 88.98\%   & 56.60\%    & 75.50\%                & no                       &                        \\
20              & 500     & unet       & 56.82\%   & 56.96\% & 71.43\% & 44.38\%   & 51.69\%  & 56.60\% & 53.38\%         & 56.82\%           & 43.04\%     & 71.43\%    & 55.63\%           & 51.69\%    & 43.40\%     & 53.38\%                 & 68.18\%          & 43.04\%    & 71.43\%   & 69.38\%          & 85.59\%   & 84.91\%    & 70.57\%                & \multicolumn{2}{l}{RonP}             \\
20              & 500     & unet       & 69.32\%   & 72.15\% & 71.43\% & 73.75\%   & 86.44\%  & 84.91\% & 76.42\%         & 69.32\%           & 70.89\%     & 71.43\%    & 73.75\%           & 86.44\%    & 84.91\%     & 76.23\%                 & 65.91\%          & 70.89\%    & 71.43\%   & 73.75\%          & 85.59\%   & 84.91\%    & 75.50\%                & \multicolumn{2}{l}{RonP}             \\
100             & 500     & unet       & 63.64\%   & 79.75\% & 73.47\% & 80.63\%   & 93.22\%  & 88.68\% & 80.62\%         & 67.05\%           & 77.22\%     & 73.47\%    & 80.63\%           & 94.07\%    & 88.68\%     & 80.99\%                 & 62.50\%          & 77.22\%    & 73.47\%   & 81.88\%          & 93.22\%   & 90.57\%    & 80.62\%                & \multicolumn{2}{l}{RonP}             \\
100             & 500     & unet       & 71.59\%   & 79.75\% & 73.47\% & 80.00\%   & 92.37\%  & 88.68\% & 81.54\%         & 72.73\%           & 79.75\%     & 77.55\%    & 79.38\%           & 93.22\%    & 88.68\%     & 82.08\%                 & 70.45\%          & 73.42\%    & 75.51\%   & 80.00\%          & 91.53\%   & 88.68\%    & 80.44\%                & no                       &                        \\
100             & 500     & no         & 68.18\%   & 78.48\% & 77.55\% & 79.38\%   & 91.53\%  & 88.68\% & 80.80\%         & 68.18\%           & 75.95\%     & 75.51\%    & 79.38\%           & 91.53\%    & 88.68\%     & 80.26\%                 & 63.64\%          & 75.95\%    & 73.47\%   & 80.00\%          & 91.53\%   & 88.68\%    & 79.52\%                & \multicolumn{2}{l}{RonP}             \\
100             & 500     & no         & 64.77\%   & 72.15\% & 75.51\% & 81.25\%   & 93.22\%  & 88.68\% & 80.07\%         & 72.73\%           & 78.48\%     & 79.59\%    & 80.63\%           & 92.37\%    & 88.68\%     & 82.27\%                 & 67.05\%          & 78.48\%    & 77.55\%   & 80.63\%          & 93.22\%   & 88.68\%    & 81.35\%                & no                       &                        \\
200             & 500     & unet       & 71.59\%   & 81.01\% & 77.55\% & 81.25\%   & 95.76\%  & 90.57\% & 83.36\%         & 69.32\%           & 79.75\%     & 77.55\%    & 80.63\%           & 96.61\%    & 92.45\%     & 83.00\%                 & 61.36\%          & 79.75\%    & 75.51\%   & 80.63\%          & 96.61\%   & 92.45\%    & 81.54\%                & \multicolumn{2}{l}{RonP}             \\
200             & 500     & unet       & 67.05\%   & 78.48\% & 79.59\% & 76.88\%   & 94.92\%  & 92.45\% & 81.17\%         & 63.64\%           & 70.89\%     & 77.55\%    & 81.25\%           & 94.92\%    & 90.57\%     & 80.44\%                 & 63.64\%          & 77.22\%    & 77.55\%   & 82.50\%          & 95.76\%   & 90.57\%    & 81.90\%                & no                       &                        \\
200             & 500     & no         & 73.86\%   & 79.75\% & 79.59\% & 81.88\%   & 95.76\%  & 90.57\% & 83.91\%         & 72.73\%           & 79.75\%     & 77.55\%    & 78.75\%           & 96.61\%    & 90.57\%     & 82.82\%                 & 65.91\%          & 78.48\%    & 81.63\%   & 78.13\%          & 96.61\%   & 92.45\%    & 81.90\%                & \multicolumn{2}{l}{RonP}             \\
200             & 500     & no         & 59.09\%   & 72.15\% & 77.55\% & 80.63\%   & 94.92\%  & 92.45\% & 79.89\%         & 67.05\%           & 74.68\%     & 79.59\%    & 81.88\%           & 96.61\%    & 92.45\%     & 82.45\%                 & 62.50\%          & 78.48\%    & 79.59\%   & 81.88\%          & 95.76\%   & 92.45\%    & 82.08\%                & no                       &                        \\
400             & 500     & unet       & 69.32\%   & 72.15\% & 79.59\% & 70.00\%   & 97.46\%  & 92.45\% & 79.16\%         & 68.18\%           & 72.15\%     & 83.67\%    & 80.00\%           & 97.46\%    & 92.45\%     & 82.27\%                 & 72.73\%          & 70.89\%    & 77.55\%   & 79.38\%          & 96.61\%   & 92.45\%    & 81.90\%                & \multicolumn{2}{l}{RonP}             \\
400             & 500     & unet       & 62.50\%   & 75.95\% & 73.47\% & 79.38\%   & 98.31\%  & 94.34\% & 81.17\%         & 67.05\%           & 77.22\%     & 73.47\%    & 76.25\%           & 95.76\%    & 94.34\%     & 80.62\%                 & 72.73\%          & 74.68\%    & 73.47\%   & 81.25\%          & 94.92\%   & 94.34\%    & 82.45\%                & no                       &                        \\
400             & 500     & no         & 56.82\%   & 73.42\% & 77.55\% & 79.38\%   & 95.76\%  & 92.45\% & 79.52\%         & 72.73\%           & 73.42\%     & 77.55\%    & 80.00\%           & 95.76\%    & 92.45\%     & 82.27\%                 & 73.86\%          & 70.89\%    & 73.47\%   & 79.38\%          & 92.37\%   & 92.45\%    & 80.80\%                & \multicolumn{2}{l}{RonP}             \\
20              & 500     & unet       & 71.59\%   & 67.09\% & 71.43\% & 80.63\%   & 83.90\%  & 75.47\% & 76.60\%         & 71.59\%           & 65.82\%     & 71.43\%    & 80.63\%           & 83.90\%    & 75.47\%     & 76.42\%                 & 62.50\%          & 65.82\%    & 71.43\%   & 80.63\%          & 78.81\%   & 77.36\%    & 74.04\%                & \multicolumn{2}{l}{RonP}             \\
20              & 100     & unet       & 56.82\%   & 56.96\% & 28.57\% & 70.63\%   & 81.36\%  & 43.40\% & 62.34\%         & 56.82\%           & 56.96\%     & 28.57\%    & 72.50\%           & 83.90\%    & 43.40\%     & 63.44\%                 & 56.82\%          & 56.96\%    & 28.57\%   & 72.50\%          & 83.05\%   & 43.40\%    & 63.25\%                & \multicolumn{2}{l}{RonP}             \\
20              & 100     & no         & 56.82\%   & 43.04\% & 71.43\% & 44.38\%   & 83.90\%  & 81.13\% & 60.69\%         & 56.82\%           & 43.04\%     & 71.43\%    & 44.38\%           & 83.90\%    & 81.13\%     & 60.69\%                 & 56.82\%          & 43.04\%    & 71.43\%   & 44.38\%          & 83.90\%   & 81.13\%    & 60.69\%                & no                       &                        \\
20              & 100     & no         & 68.18\%   & 74.68\% & 71.43\% & 70.00\%   & 85.59\%  & 73.58\% & 74.22\%         & 68.18\%           & 74.68\%     & 71.43\%    & 72.50\%           & 85.59\%    & 73.58\%     & 74.95\%                 & 68.18\%          & 74.68\%    & 71.43\%   & 71.88\%          & 85.59\%   & 71.70\%    & 74.59\%                & no                       &                        \\
20              & 100     & no         & 64.77\%   & 73.42\% & 71.43\% & 73.75\%   & 76.27\%  & 69.81\% & 72.21\%         & 69.32\%           & 73.42\%     & 71.43\%    & 75.00\%           & 75.42\%    & 69.81\%     & 73.13\%                 & 69.32\%          & 72.15\%    & 71.43\%   & 73.75\%          & 83.05\%   & 75.47\%    & 74.77\%                & no                       &                        \\
20              & 100     & no         & 56.82\%   & 43.04\% & 71.43\% & 71.25\%   & 83.90\%  & 77.36\% & 68.19\%         & 56.82\%           & 43.04\%     & 71.43\%    & 69.38\%           & 84.75\%    & 73.58\%     & 67.46\%                 & 59.09\%          & 43.04\%    & 71.43\%   & 71.25\%          & 83.90\%   & 75.47\%    & 68.37\%                & no                       &                        \\
20              & 100     & VAE        & 64.77\%   & 69.62\% & 71.43\% & 75.00\%   & 83.90\%  & 77.36\% & 74.41\%         & 67.05\%           & 69.62\%     & 71.43\%    & 74.38\%           & 84.75\%    & 79.25\%     & 74.95\%                 & 68.18\%          & 70.89\%    & 71.43\%   & 73.75\%          & 83.90\%   & 81.13\%    & 75.14\%                & no                       &                        \\
20              & 100     & VAE        & 68.18\%   & 64.56\% & 71.43\% & 72.50\%   & 86.44\%  & 79.25\% & 74.22\%         & 68.18\%           & 70.89\%     & 71.43\%    & 72.50\%           & 85.59\%    & 79.25\%     & 74.95\%                 & 62.50\%          & 67.09\%    & 71.43\%   & 68.13\%          & 85.59\%   & 77.36\%    & 72.03\%                & no                       &                        \\
20              & 500     & VAE        & 72.73\%   & 68.35\% & 77.55\% & 76.25\%   & 82.20\%  & 75.47\% & 75.87\%         & 76.14\%           & 60.76\%     & 77.55\%    & 80.00\%           & 83.05\%    & 71.70\%     & 76.23\%                 & 71.59\%          & 62.03\%    & 79.59\%   & 76.25\%          & 82.20\%   & 75.47\%    & 74.95\%                & no                       &                        \\
20              & 500     & VAE        & 75.00\%   & 72.15\% & 83.67\% & 81.25\%   & 83.05\%  & 77.36\% & 79.16\%         & 72.73\%           & 75.95\%     & 83.67\%    & 81.25\%           & 83.05\%    & 77.36\%     & 79.34\%                 & 73.86\%          & 77.22\%    & 85.71\%   & 79.38\%          & 84.75\%   & 77.36\%    & 79.71\%                & \multicolumn{2}{l}{RonP}             \\
20              & 5       & V\&u  & 56.82\%   & 56.96\% & 28.57\% & 44.38\%   & 51.69\%  & 43.40\% & 48.26\%         & 56.82\%           & 56.96\%     & 28.57\%    & 44.38\%           & 51.69\%    & 43.40\%     & 48.26\%                 & 56.82\%          & 56.96\%    & 28.57\%   & 44.38\%          & 51.69\%   & 43.40\%    & 48.26\%                & \multicolumn{2}{l}{RonP}             \\
100             & 500     & VAE        & 65.91\%   & 79.75\% & 69.39\% & 78.75\%   & 94.07\%  & 92.45\% & 80.62\%         & 68.18\%           & 78.48\%     & 69.39\%    & 79.38\%           & 93.22\%    & 90.57\%     & 80.62\%                 & 64.77\%          & 78.48\%    & 69.39\%   & 80.63\%          & 91.53\%   & 90.57\%    & 80.07\%                & \multicolumn{2}{l}{RonP}             \\
100             & 500     & VAE        & 78.41\%   & 82.28\% & 79.59\% & 80.00\%   & 92.37\%  & 90.57\% & 83.73\%         & 69.32\%           & 81.01\%     & 77.55\%    & 80.63\%           & 91.53\%    & 86.79\%     & 81.54\%                 & 64.77\%          & 82.28\%    & 79.59\%   & 82.50\%          & 91.53\%   & 86.79\%    & 81.72\%                & no                       &                        \\
100             & 500     & V\&u  & 72.73\%   & 83.54\% & 73.47\% & 79.38\%   & 92.37\%  & 88.68\% & 82.08\%         & 65.91\%           & 83.54\%     & 73.47\%    & 81.25\%           & 93.22\%    & 86.79\%     & 81.54\%                 & 68.18\%          & 83.54\%    & 71.43\%   & 81.88\%          & 93.22\%   & 88.68\%    & 82.08\%                & \multicolumn{2}{l}{RonP}             \\
100             & 500     & V\&u  & 70.45\%   & 69.62\% & 81.63\% & 80.63\%   & 92.37\%  & 92.45\% & 81.17\%         & 68.18\%           & 73.42\%     & 75.51\%    & 82.50\%           & 91.53\%    & 90.57\%     & 80.99\%                 & 62.50\%          & 82.28\%    & 71.43\%   & 76.88\%          & 93.22\%   & 92.45\%    & 79.89\%                & no                       &                        \\
100             & 500     & unet       & 67.05\%   & 83.54\% & 73.47\% & 78.75\%   & 93.22\%  & 90.57\% & 81.35\%         & 70.45\%           & 83.54\%     & 75.51\%    & 79.38\%           & 93.22\%    & 90.57\%     & 82.27\%                 & 78.41\%          & 65.82\%    & 71.43\%   & 77.50\%          & 92.37\%   & 90.57\%    & 79.89\%                & \multicolumn{2}{l}{RonP}             \\
100             & 500     & unet       & 62.50\%   & 69.62\% & 77.55\% & 79.38\%   & 93.22\%  & 88.68\% & 78.98\%         & 72.73\%           & 78.48\%     & 79.59\%    & 78.75\%           & 93.22\%    & 88.68\%     & 81.90\%                 & 65.91\%          & 78.48\%    & 75.51\%   & 79.38\%          & 92.37\%   & 86.79\%    & 80.26\%                & no                       &                        \\
200             & 500     & VAE        & 70.45\%   & 75.95\% & 73.47\% & 82.50\%   & 95.76\%  & 88.68\% & 82.27\%         & 67.05\%           & 75.95\%     & 73.47\%    & 82.50\%           & 97.46\%    & 92.45\%     & 82.45\%                 & 62.50\%          & 77.22\%    & 77.55\%   & 79.38\%          & 97.46\%   & 92.45\%    & 81.35\%                & \multicolumn{2}{l}{RonP}             \\
200             & 500     & VAE        & 69.32\%   & 78.48\% & 73.47\% & 81.25\%   & 93.22\%  & 90.57\% & 81.72\%         & 69.32\%           & 75.95\%     & 81.63\%    & 79.38\%           & 94.92\%    & 92.45\%     & 82.08\%                 & 78.41\%          & 81.01\%    & 79.59\%   & 80.00\%          & 94.07\%   & 92.45\%    & 84.10\%                & no                       &                        \\
200             & 500     & V\&u  & 69.32\%   & 79.75\% & 79.59\% & 78.13\%   & 96.61\%  & 90.57\% & 82.27\%         & 70.45\%           & 77.22\%     & 79.59\%    & 81.88\%           & 97.46\%    & 90.57\%     & 83.36\%                 & 73.86\%          & 73.42\%    & 73.47\%   & 78.75\%          & 94.92\%   & 90.57\%    & 81.35\%                & \multicolumn{2}{l}{RonP}             \\
200             & 500     & V\&u  & 78.41\%   & 74.68\% & 73.47\% & 73.13\%   & 94.92\%  & 88.68\% & 80.44\%         & 71.59\%           & 73.42\%     & 73.47\%    & 81.88\%           & 96.61\%    & 92.45\%     & 82.45\%                 & 65.91\%          & 78.48\%    & 71.43\%   & 80.63\%          & 97.46\%   & 88.68\%    & 81.54\%                & no                       &                        \\
200             & 500     & unet       & 65.91\%   & 79.75\% & 79.59\% & 78.75\%   & 96.61\%  & 90.57\% & 81.90\%         & 69.32\%           & 77.22\%     & 77.55\%    & 81.88\%           & 98.31\%    & 88.68\%     & 83.00\%                 & 65.91\%          & 79.75\%    & 77.55\%   & 77.50\%          & 97.46\%   & 88.68\%    & 81.35\%                & \multicolumn{2}{l}{RonP}             \\
200             & 500     & unet       & 73.86\%   & 73.42\% & 77.55\% & 81.25\%   & 96.61\%  & 88.68\% & 82.63\%         & 68.18\%           & 73.42\%     & 75.51\%    & 82.50\%           & 96.61\%    & 90.57\%     & 82.08\%                 & 67.05\%          & 73.42\%    & 75.51\%   & 81.25\%          & 94.07\%   & 90.57\%    & 80.99\%                & no                       &                        \\
400             & 500     & VAE        & 56.82\%   & 64.56\% & 77.55\% & 81.88\%   & 94.92\%  & 94.34\% & 78.98\%         & 71.59\%           & 74.68\%     & 77.55\%    & 80.00\%           & 94.92\%    & 92.45\%     & 82.08\%                 & 73.86\%          & 73.42\%    & 77.55\%   & 79.38\%          & 99.15\%   & 92.45\%    & 83.00\%                & \multicolumn{2}{l}{RonP}             \\
400             & 500     & VAE        & 55.68\%   & 75.95\% & 75.51\% & 80.00\%   & 94.92\%  & 88.68\% & 79.16\%         & 65.91\%           & 73.42\%     & 75.51\%    & 79.38\%           & 96.61\%    & 88.68\%     & 80.62\%                 & 72.73\%          & 70.89\%    & 77.55\%   & 80.00\%          & 95.76\%   & 86.79\%    & 81.35\%                & no                       &                        \\
400             & 500     & V\&u  & 71.59\%   & 78.48\% & 75.51\% & 80.63\%   & 96.61\%  & 86.79\% & 82.45\%         & 65.91\%           & 77.22\%     & 75.51\%    & 78.75\%           & 96.61\%    & 90.57\%     & 81.17\%                 & 64.77\%          & 65.82\%    & 81.63\%   & 78.75\%          & 97.46\%   & 92.45\%    & 80.26\%                & \multicolumn{2}{l}{RonP}             \\
400             & 500     & V\&u  & 65.91\%   & 77.22\% & 71.43\% & 81.25\%   & 94.07\%  & 43.40\% & 76.42\%         & 71.59\%           & 74.68\%     & 71.43\%    & 80.00\%           & 93.22\%    & 43.40\%     & 76.42\%                 & 64.77\%          & 74.68\%    & 77.55\%   & 81.25\%          & 94.07\%   & 43.40\%    & 76.42\%                & no                       &                        \\
400             & 500     & unet       & 63.64\%   & 78.48\% & 79.59\% & 80.00\%   & 96.61\%  & 43.40\% & 77.15\%         & 71.59\%           & 72.15\%     & 81.63\%    & 80.00\%           & 97.46\%    & 43.40\%     & 77.88\%                 & 61.36\%          & 78.48\%    & 81.63\%   & 79.38\%          & 98.31\%   & 43.40\%    & 77.15\%                & \multicolumn{2}{l}{RonP}             \\
400             & 500     & unet       & 59.09\%   & 74.68\% & 81.63\% & 72.50\%   & 95.76\%  & 94.34\% & 78.61\%         & 64.77\%           & 72.15\%     & 81.63\%    & 80.00\%           & 97.46\%    & 94.34\%     & 81.72\%                 & 72.73\%          & 70.89\%    & 81.63\%   & 80.63\%          & 98.31\%   & 94.34\%    & 83.18\%                & no                       &                        \\
1000            & 500     & VAE        & 43.18\%   & 70.89\% & 73.47\% & 80.00\%   & 100.00\% & 90.57\% & 77.51\%         & 68.18\%           & 70.89\%     & 73.47\%    & 78.75\%           & 100.00\%   & 92.45\%     & 81.35\%                 & 70.45\%          & 69.62\%    & 71.43\%   & 75.63\%          & 98.31\%   & 92.45\%    & 80.07\%                & \multicolumn{2}{l}{RonP}             \\
1000            & 500     & VAE        & 71.59\%   & 63.29\% & 71.43\% & 56.25\%   & 51.69\%  & 88.68\% & 63.25\%         & 68.18\%           & 63.29\%     & 77.55\%    & 57.50\%           & 79.66\%    & 90.57\%     & 69.84\%                 & 70.45\%          & 58.23\%    & 75.51\%   & 67.50\%          & 64.41\%   & 90.57\%    & 68.92\%                & no                       &                        \\
1000            & 500     & V\&u  & 69.32\%   & 70.89\% & 79.59\% & 79.38\%   & 94.92\%  & 88.68\% & 80.80\%         & 65.91\%           & 69.62\%     & 77.55\%    & 78.75\%           & 94.92\%    & 88.68\%     & 79.71\%                 & 63.64\%          & 79.75\%    & 77.55\%   & 80.63\%          & 91.53\%   & 90.57\%    & 80.80\%                & \multicolumn{2}{l}{RonP}             \\
1000            & 500     & V\&u  & 68.18\%   & 70.89\% & 75.51\% & 79.38\%   & 99.15\%  & 94.34\% & 81.72\%         & 68.18\%           & 70.89\%     & 75.51\%    & 83.13\%           & 99.15\%    & 94.34\%     & 82.82\%                 & 75.00\%          & 58.23\%    & 77.55\%   & 77.50\%          & 98.31\%   & 94.34\%    & 80.44\%                & no                       &                        \\
1000            & 500     & unet       & 65.91\%   & 68.35\% & 73.47\% & 78.13\%   & 83.90\%  & 92.45\% & 76.97\%         & 69.32\%           & 68.35\%     & 75.51\%    & 79.38\%           & 96.61\%    & 90.57\%     & 80.62\%                 & 72.73\%          & 70.89\%    & 75.51\%   & 78.13\%          & 97.46\%   & 88.68\%    & 81.17\%                & \multicolumn{2}{l}{RonP}             \\
1000            & 500     & unet       & 72.73\%   & 65.82\% & 73.47\% & 59.38\%   & 54.24\%  & 90.57\% & 65.63\%         & 68.18\%           & 73.42\%     & 71.43\%    & 79.38\%           & 91.53\%    & 90.57\%     & 79.71\%                 & 70.45\%          & 68.35\%    & 75.51\%   & 76.25\%          & 94.92\%   & 92.45\%    & 79.71\%                & no                       &                        \\
2000            & 500     & V\&u  & 73.86\%   & 73.42\% & 77.55\% & 78.13\%   & 98.31\%  & 90.57\% & 82.27\%         & 70.45\%           & 72.15\%     & 77.55\%    & 81.25\%           & 99.15\%    & 90.57\%     & 82.63\%                 & 63.64\%          & 73.42\%    & 77.55\%   & 80.63\%          & 99.15\%   & 90.57\%    & 81.54\%                & \multicolumn{2}{l}{RonP}             \\
2000            & 500     & V\&u  & 68.18\%   & 50.63\% & 71.43\% & 80.00\%   & 58.47\%  & 88.68\% & 69.29\%         & 69.32\%           & 69.62\%     & 71.43\%    & 83.75\%           & 69.49\%    & 90.57\%     & 75.87\%                 & 65.91\%          & 70.89\%    & 71.43\%   & 77.50\%          & 84.75\%   & 90.57\%    & 76.97\%                & no                       &                        \\
2000            & 500     & unet       & 75.00\%   & 63.29\% & 71.43\% & 78.13\%   & 69.49\%  & 90.57\% & 74.22\%         & 67.05\%           & 67.09\%     & 79.59\%    & 77.50\%           & 72.88\%    & 90.57\%     & 74.77\%                 & 71.59\%          & 77.22\%    & 77.55\%   & 80.00\%          & 76.27\%   & 92.45\%    & 78.43\%                & \multicolumn{2}{l}{RonP}             \\
2000            & 500     & unet       & 72.73\%   & 70.89\% & 75.51\% & 76.88\%   & 99.15\%  & 88.68\% & 81.17\%         & 67.05\%           & 73.42\%     & 85.71\%    & 78.75\%           & 99.15\%    & 90.57\%     & 82.27\%                 & 60.23\%          & 72.15\%    & 71.43\%   & 81.88\%          & 100.00\%  & 88.68\%    & 80.62\%                & no                       &                        \\
20              & 500     & V\&u  & 76.14\%   & 73.42\% & 79.59\% & 79.38\%   & 83.90\%  & 71.70\% & 78.24\%         & 76.14\%           & 73.42\%     & 79.59\%    & 78.75\%           & 83.05\%    & 71.70\%     & 77.88\%                 & 76.14\%          & 74.68\%    & 75.51\%   & 79.38\%          & 83.05\%   & 71.70\%    & 77.88\%                & \multicolumn{2}{l}{RonP}             \\
20              & 500     & V\&u  & 77.27\%   & 75.95\% & 81.63\% & 79.38\%   & 82.20\%  & 86.79\% & 80.07\%         & 76.14\%           & 75.95\%     & 81.63\%    & 78.75\%           & 82.20\%    & 86.79\%     & 79.71\%                 & 78.41\%          & 75.95\%    & 75.51\%   & 80.63\%          & 82.20\%   & 86.79\%    & 80.07\%                & \multicolumn{2}{l}{RonP}             \\
200             & 500     & V\&u  & 69.32\%   & 72.15\% & 77.55\% & 78.75\%   & 96.61\%  & 92.45\% & 81.35\%         & 70.45\%           & 70.89\%     & 79.59\%    & 83.13\%           & 97.46\%    & 92.45\%     & 83.00\%                 & 65.91\%          & 79.75\%    & 79.59\%   & 80.63\%          & 95.76\%   & 92.45\%    & 82.45\%                & \multicolumn{2}{l}{RonP}            
\end{tabular}
\end{table}

\begin{figure}[htbp]
        \centering
        \includegraphics[width=0.8\linewidth]{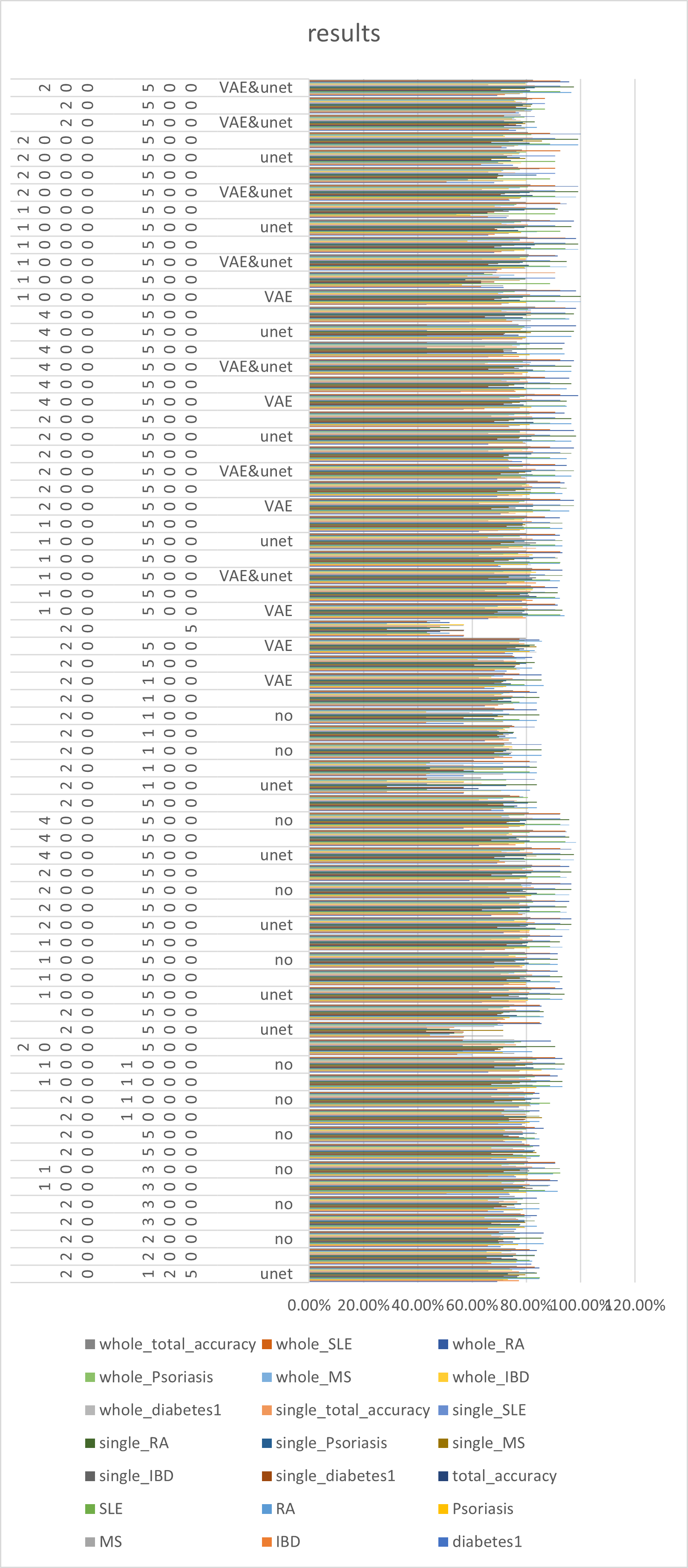}
   \caption{experimental results}
\end{figure}
\begin{figure}[htbp]
        \centering
        \includegraphics[width=1\linewidth]{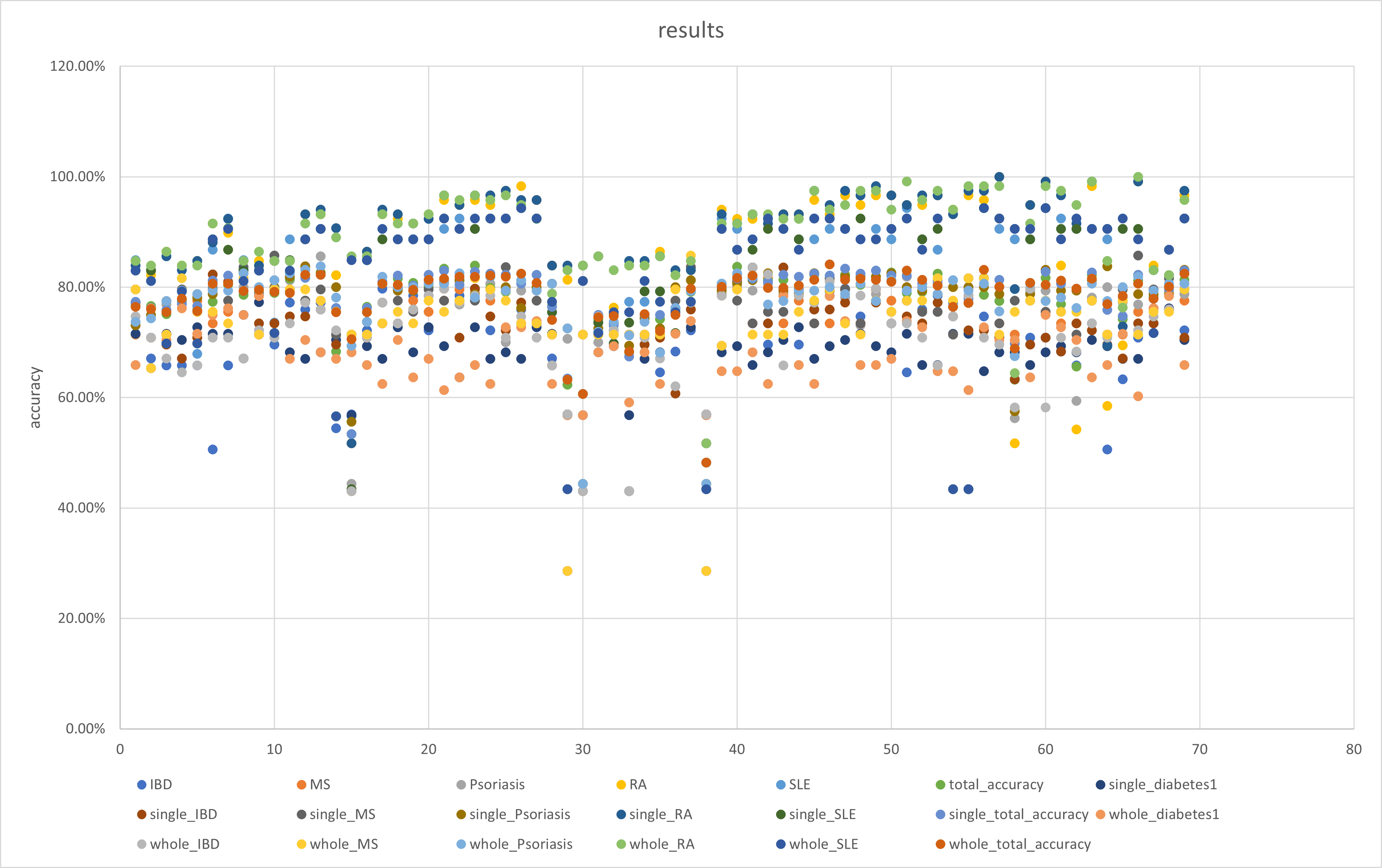}
   \caption{experimental results}
\end{figure}

In dataset accuracy: "diab1" means "type 1 diabetes", "Ps"means "Psoriasis".
In model column,"no" means normal autoencoder architecture,"unet" means Autoencoder+unet architecture, $V\&u$ means VAE+unet architecture, .
In train policy column, "RonP" means ReduceLROnPlateau.

From the experiments, we observe that the performance of the models varies across different configurations. The best-performing models achieve an accuracy of over 80\% in some cases, while other configurations result in accuracies as low as 50\% or 60\%. The use of a multi-task learning policy, such as RonP, leads to improvements in some cases, but not consistently across all configurations.
In conclusion, the paper explores the effectiveness of methylation embedding using interpretable neural networks for the prediction of various phenotypes. The results indicate that careful selection of model configurations and multi-task learning policies is essential for achieving optimal performance.

For Plot 1, we observe that as the number of selected sites increases, the accuracy of the model generally improves. This indicates that incorporating more sites in the model allows it to better capture and generalize the underlying patterns in the data. However, beyond a certain point, the improvement in accuracy plateaus, suggesting that there may be a limit to how much additional information can be gained from including more sites.
For Plot 2, we examine the relationship between the number of training epochs and the model's accuracy for three different numbers of selected sites (25, 200, and 2000). In each case, we observe an initial rapid increase in accuracy as the number of epochs grows. This suggests that the model is effectively learning from the data during the early stages of training.
As the number of epochs increases further, the rate of improvement in accuracy begins to slow down. Eventually, the accuracy curve starts to plateau, indicating that the model has reached its capacity for learning from the given dataset. Comparing the three separate plots, it is evident that the model trained on a larger number of sites (2000) achieves higher accuracy levels than those with fewer sites (25 and 200), reinforcing the findings from Plot 1.
\newpage
\section{Weight Results}
\begin{figure}[H]
        \centering
        \includegraphics[width=1\linewidth]{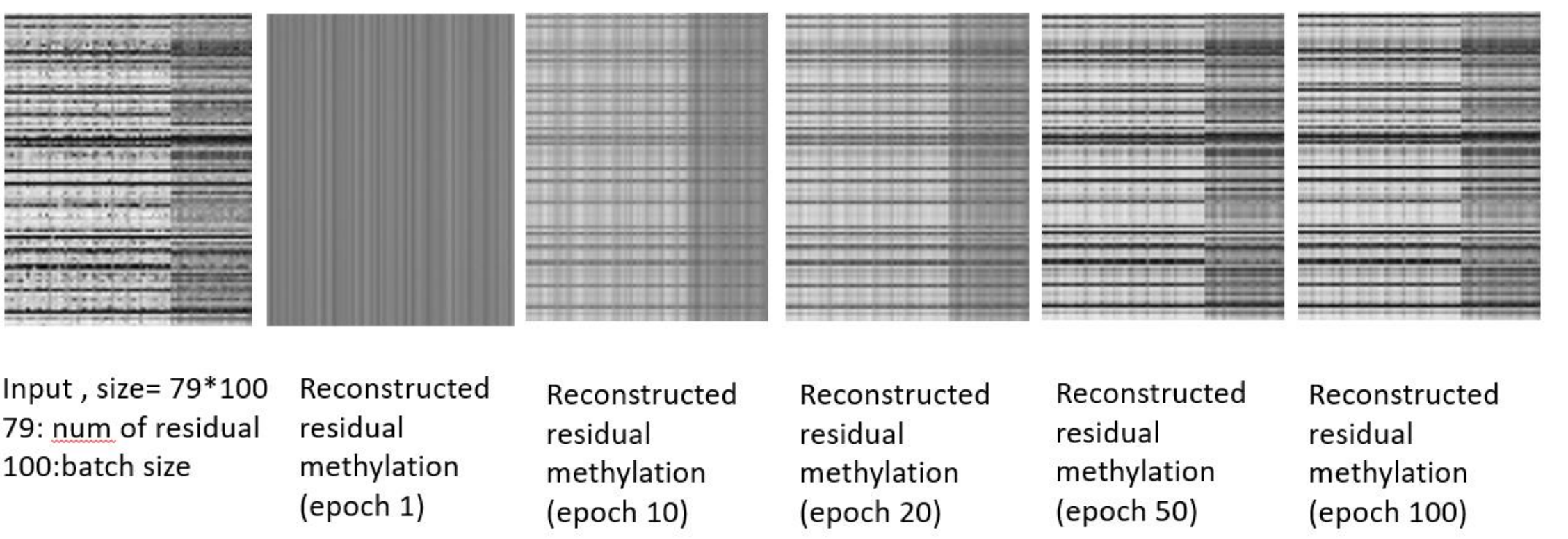}
   \caption{Input and AutoEncoder reconstructed input after different epochs}
\end{figure}
\begin{figure}[H]
        \centering
        \includegraphics[width=0.8\linewidth]{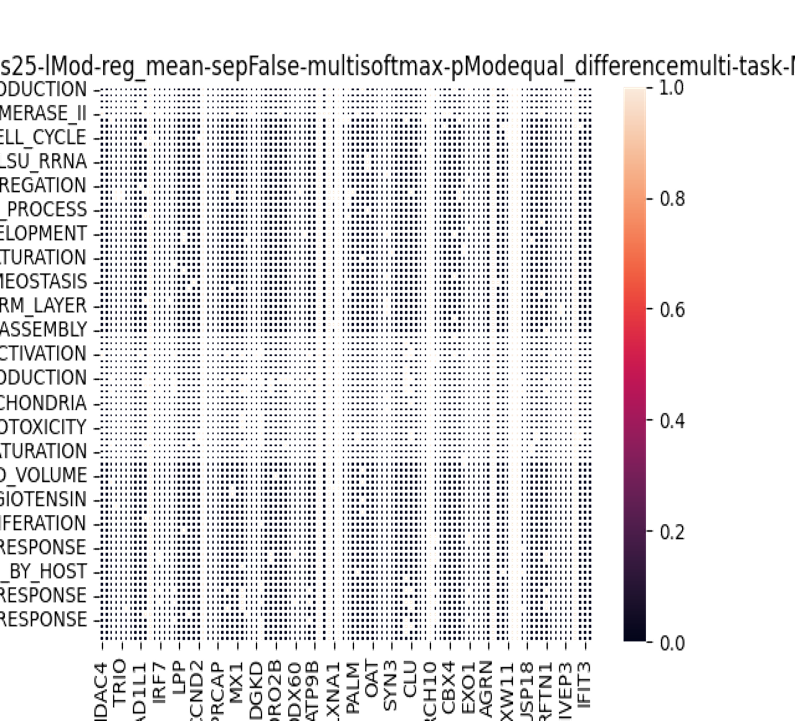}
   \caption{Adjacency matrix of gene-pathway }
\end{figure}
\begin{figure}[H]
        \centering
        \includegraphics[width=0.8\linewidth]{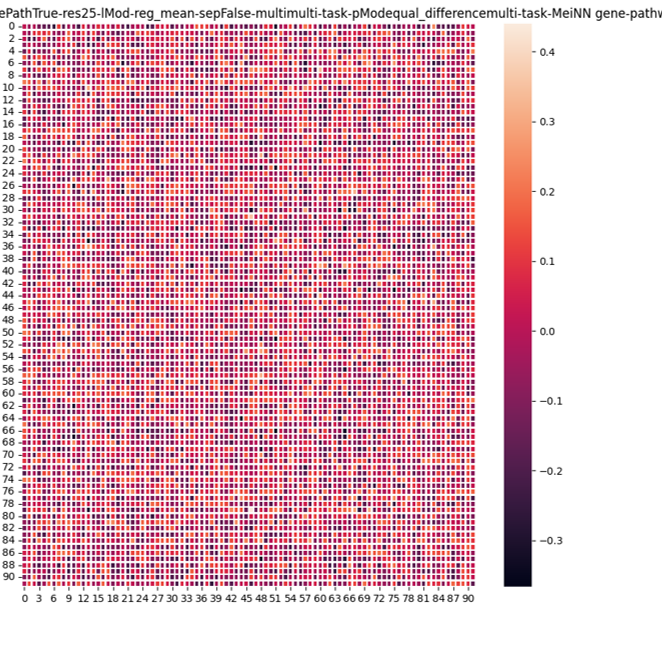}
   \caption{HeatMap of gene-pathway layer}
\end{figure}
\begin{figure}[H]
        \centering
        \includegraphics[width=0.8\linewidth]{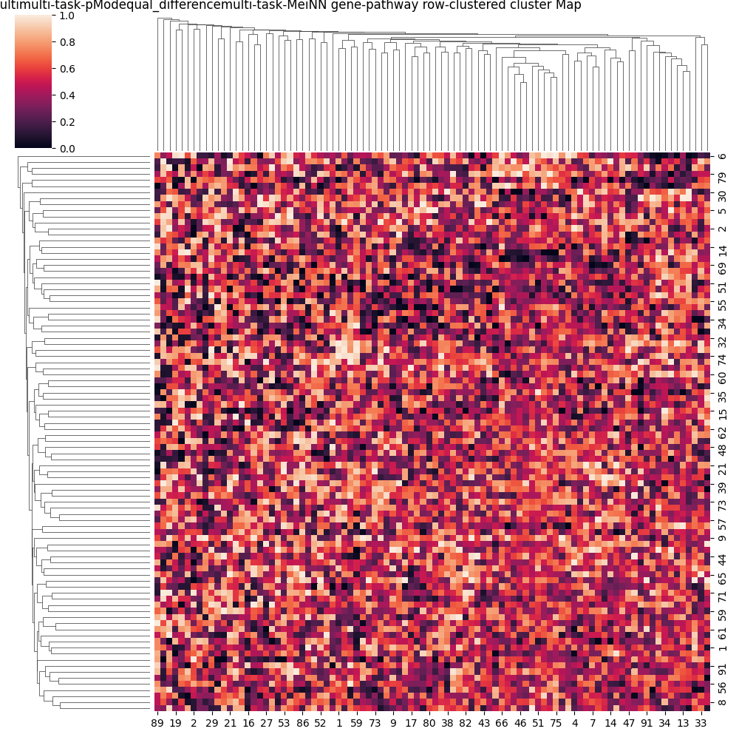}
   \caption{clustermap of gene-pathway layer}
\end{figure}

\newpage

% References follow the acknowledgments in the camera-ready paper. Use unnumbered first-level heading for
% the references. Any choice of citation style is acceptable as long as you are
% consistent. It is permissible to reduce the font size to \verb+small+ (9 point)
% when listing the references.
% Note that the Reference section does not count towards the page limit.
\medskip

\end{document}